\documentclass{article}

\usepackage{microtype}
\usepackage{graphicx}
\usepackage{subfigure}
\usepackage{booktabs} 

\usepackage{hyperref}

\usepackage[accepted]{icml2023}

\usepackage{amsmath}
\usepackage{amssymb}
\usepackage{mathtools}
\usepackage{amsthm}

\usepackage[capitalize,noabbrev]{cleveref}

\newcommand{\bs}[1]{\boldsymbol{#1}}
\newcommand{\mbf}[1]{\mathbf{#1}}
\newcommand{\mbb}[1]{\mathbb{#1}}
\newcommand{\mca}[1]{\mathcal{#1}}

\newcommand{\dev}{\mathrm{d}}
\newcommand{\indep}{\perp \!\!\! \perp}

\theoremstyle{plain}
\newtheorem{theorem}{Theorem}[section]
\newtheorem{proposition}[theorem]{Proposition}
\newtheorem{lemma}[theorem]{Lemma}

\theoremstyle{definition}

\theoremstyle{remark}

\usepackage[textsize=tiny]{todonotes}

\icmltitlerunning{Adversarially Contrastive Estimation of Conditional Neural Processes}

\begin{document}

\twocolumn[
\icmltitle{Adversarially Contrastive Estimation of Conditional Neural Processes}



\icmlsetsymbol{equal}{*}

\begin{icmlauthorlist}
\icmlauthor{Zesheng Ye}{xxx}
\icmlauthor{Jing Du}{xxx}
\icmlauthor{Lina Yao}{xxx,yyy}
\end{icmlauthorlist}

\icmlaffiliation{xxx}{The University of New South Wales}
\icmlaffiliation{yyy}{CSIRO's Data61}

\icmlcorrespondingauthor{Zesheng Ye}{zesheng.ye@unsw.edu.au}

\icmlkeywords{Neural Processes, Noise Contrastive Estimation, Adversarial Training}

\vskip 0.3in
]



\printAffiliationsAndNotice{}  

\begin{abstract}

Conditional Neural Processes~(CNPs) formulate distributions over functions and generate function observations with exact conditional likelihoods.
CNPs, however, have limited expressivity for high-dimensional observations, since their predictive distribution is factorized into a product of unconstrained (typically) Gaussian outputs.
Previously, this could be handled using latent variables or autoregressive likelihood, but at the expense of intractable training and quadratically increased complexity.
Instead, we propose calibrating CNPs with an adversarial training scheme besides regular maximum likelihood estimates.
Specifically, we train an energy-based model (EBM) with noise contrastive estimation, which enforces EBM to identify true observations from the generations of CNP.
In this way, CNP must generate predictions closer to the ground-truth to fool EBM, instead of merely optimizing with respect to the fixed-form likelihood.
From generative function reconstruction to downstream regression and classification tasks, we demonstrate that our method fits mainstream CNP members, showing effectiveness when unconstrained Gaussian likelihood is defined, requiring minimal computation overhead while preserving foundation properties of CNPs.

\end{abstract}

\section{Introduction}
\label{sec: intro}

Conditional Neural Processes~(CNPs)~\cite{garnelo2018conditional} are a class of neural density estimators derived from function approximation in meta-learning setting~\cite{hospedales2021meta}.
Basically, CNPs learn an empirical prior from seen function instantiations, allowing quick adaptation to new functions within a single forward pass.
Dividing each instantiation into a {\it context} and a {\it target} set,
the prior is concluded as a {\it context} representation like mean function in Gaussian Processes~\cite{williams2006gaussian}, but is parameterized by a neural network-based {\it encoder}.
Using this prior, {\it decoder} generates function observations $\bs{y}_t$ by parameterizing a predictive distribution over all the {\it target} function covariates $\bs{x}_t$.
Namely, CNPs are trained by exact maximum likelihood estimation~(MLE).
This endows CNPs with desirable properties such as tractable density and quantifiable uncertainty, suggesting them a good fit for image impainting~\cite{kimanp18}, transfer learning~\cite{Gordon2019ConvCNP} and climate modeling~\cite{holderrieth2021equivariant}.

Tractability is indeed important, but it can also be a double-edged sword sometimes.
In CNPs, the predictive distribution is conditionally factorized, which strips away correlations {\it between} samples, challenging their ability to generate coherent {\it target} observations from the predictive space~\cite{garnelo2018neural}.
Moreover, CNPs specify a fixed likelihood function (usually Gaussian) and assume true observations can be sampled exactly from such generative processes.
The expressivity of models defined in this way is challenged when observations spanning high-dimensional spaces~\cite{rybkin2021simple}.
Often, unconstrained Gaussian outputs fail to correctly learn correlations between dimensions and variances {\it within} each observation, leading to ill-posed MLE training~\cite{rezende2018taming, mattei2018leveraging}.

Prior works opened up two paths in addressing the issues above.
One started from latent Neural Processes~(NPs)~\cite{garnelo2018neural} that introduce a latent variable to enrich the predictive space.
Following, the single latent variable can be extended to hierarchical NPs to incorporate example-specific stochasticity~\cite{wang2020doubly}, combine function spaces of multiple tasks~\cite{kim2021multi} and induce a mixture of NPs~\cite{wang2022learning} to handle complicated scenarios.
The latent variables, however, have intractable integrals and need to be estimated by a variational lower bound~\cite{garnelo2018neural} or Monte-Carlo approximations~\cite{Foong2020ConvNP} instead.
Even so, approximation does not guarantee the quality of latent representation~\cite{alemi2018fixing}.
Recently, Transformer Neural Processes~(TNP)~\cite{nguyen2022transformer} flags another workaround that replaces the independent and simultaneous {\it target} predictions with an autoregressive likelihood.
TNP trades off the foundational prediction consistency of CNPs in exchange for improved performance.
Further, the integrated casual mask attention~\cite{brown2020language} incurs quadratic computational costs as the {\it context} size grows.

Alternatively, noise contrastive estimation~(NCE)~\cite{gutmann2010noise} has gained popularity in representation learning with complex distributions~\cite{oord2018representation}.
Using a density ratio trick, NCE bypasses direct density estimation by learning the difference between true distribution and noise.
This proves to be effective in tasks involving high-dimensional data~\cite{chen2020simple, radford2021learning}.
Previously combined with CNPs, NCE creates robust meta-representations across functions by aligning partial views of the same function and contrasting them with other functions~\cite{gondal2021function}.
By using self-attention~\cite{vaswani2017attention}, distances between covariates can further be encoded into the {\it context} representation, so as to better account for their correlations~\cite{mathieu2021contrastive}.
Still, they primarily focus on learning a transferable {\it context} representation to downstream tasks and even remove the {\it decoder} from CNPs.
While this motivation is also applicable as a regularization~\cite{kallidromitis2021contrastive}, no principled clue is given for how merely improving the {\it context} would benefit the generative objective of CNPs~\cite{mathieu2021contrastive}.

To fill this gap, we present a noise contrastive approach to directly calibrate the CNP predictions.
Apart from the MLE objective, we provide CNPs with additional discriminative feedback evaluating if their predictions are indistinguishable from ground-truth.
This prevents CNPs from solely optimizing based on inherent assumptions about data distributions.
Specifically, we include an energy-based model~(EBMs)~\cite{lecun2006tutorial} learning to distinguish the true observations from those generated by CNPs.
The key idea is to lift the expressivity of CNPs, which is usually restricted to a fixed form of predictive likelihood.
In contrast, EBMs do not confine function form and normalized output, showing capacity to estimate complex distributions~\cite{du2019implicit}, yet are accused of hardship in exact density estimation and sampling.
Having said that, NCE suggests a principle for training EBMs to estimate an unknown distribution by discriminating between data drawn therefrom and noises from a known distribution.
In this context, CNPs are ideal noise generators provided that 1) samples can be easily obtained by viewing them as prediction maps~\cite{markou2022practical}, and 2) density function is analytically tractable and exact.

Intuitively, we can fit such a NCE-based calibration of CNP into an adversarial training framework~\cite{goodfellow2020generative}.
As the training progresses iteratively, CNP learns to generate increasingly realistic observations to avoid being identified easily, whereas EBM captures more discriminative characteristics in the ground-truth functions, which are then passed to CNP as gradients.
It equates to approximately minimizing the Jenson-Shannon Divergence between CNP's predictive distribution and true generative process if EBM reaches its optimality~\cite{gui2021review}.
In CNPs, this can be used as a regularization to complement the MLE objective.

In summary, our contributions are: 
i) we introduce NCE into the standard training of CNPs by adopting a flexible EBM to constrain CNP's generative process, where EBM guides CNPs to generate predictions closer to the true observations {\it from a discriminative viewpoint}, helping to relieve ill-posed MLE training.
ii) we propose integrating the joint training of EBM and CNP under an adversarial framework.
The mini-max game allows EBM and CNP to be promoted iteratively, {\it while preserving the merits of lightweight inferences by CNPs}. 
iii) we show that the proposed approach is plug-and-play with any CNP variant, consistently {\it improving generation} performances while also {\it facilitating downstream tasks}, evaluated with both artificial and real-world datasets.



\section{Preliminaries}
\label{sec: background}
\subsection{Conditional Neural Processes}
Conditional Neural Processes (CNPs)~\cite{garnelo2018conditional} are parametric models that approximate stochastic processes with {\it explicit} density functions in the meta-learning setting.
A stochastic process defines a family of data-generating functions $\mca{F}: \mca{X} \to \mca{Y}$ from covariate space $\mca{X}$ to an observation space $\mca{Y}$.
Let $F$ be a collection of function instantiations sampled from the prior $p(\mca{F})$.
We call $F$ a {\it meta}-dataset.
Each instantiation $f \in F$ samples a finite sequence of covariate-observation pairs $\{(\bs{x}_{n}, \bs{y}_{n})\}_{n=1:N}$ to describe a particular function mapping with noises $\xi$, such that $\mbf{y} = f(\mbf{x}; \xi) $ with $\mbf{x} \subseteq \mbb{R}^{d_{\mca{X}}}$ and $\mbf{y} \subseteq \mbb{R}^{d_{\mca{Y}}}$.
Under exchangeability and consistency required by the Kolmogorov Extension Theorem~\cite{oksendal2003stochastic}, the marginal joint distribution of these $N$ observations is as,
\begin{equation}
    p(\bs{y}_{1:N} | \bs{x}_{1:N} ) = \int p(f) p(\bs{y}_{1:N} | \bs{x}_{1:N}, f) \, \dev f.
\end{equation}
Provided a subset collection $F_{train} = \{ f_{k} \}_{k=1:K}$ called {\it meta-train} set, CNPs are tasked with inferring each underlying function.
For any $f_k$, CNPs randomly constitute a {\it context} set $D_{C}^{k} = \{(\bs{x}_{c}^{k}, \bs{y}_{c}^{k})\}_{c=1:C}$ from $N$ pairs and generate observations on the {\it target} set $D_{T}^{k} = \{ (\bs{x}_{t}^{k}, \bs{y}_{t}^{k}) \}_{t=1:T}$.
Using an empirical approximation of prior $p(f_k)$, the conditional generation of {\it target} observations is given by parameterizing a factorized predictive distribution over {\it target} covariates, with a known density likelihood~(e.g., a Gaussian).
Denote model parameters by $\theta$, CNPs entail a maximum likelihood estimation~(MLE) within all instantiations,
\begin{equation}\label{eq:cnp_fac}
    \begin{aligned}
        \max_{\theta} \; & \mbb{E}_{f_k \sim F_{train}} \left[ p \left(\bs{y}_{1:T}^{k} | \bs{x}_{1:T}^{k} ; \theta \right)\right] \\
        = & \int \prod_{t=1}^{T} p(f_k | D_{C}^{k} ; \theta) \, p(\bs{y}_t^{k} | \bs{x}_t^{k}, f; \theta) \, \dev f_k \, .
    \end{aligned}
\end{equation}
Notably, the {\it explicit} likelihood and factorized predictions ensure CNPs perform exact density estimation.

The fitted CNP model $\theta^{*}$ is then evaluated on another subset of collections called {\it meta-test} set, i.e., $F_{test} \cap F_{train} = \varnothing$.
The downstream evaluation task may or may not directly relate to model training.
For instance,~\cite{garnelo2018conditional, garnelo2018neural} assess the generation quality of $N$ {\it target} observations\footnote{Definition of {\it target} set varies across CNPs literature; a common practice is by following~\cite{le2018empirical} that considers {\it target} set a superset of the {\it context} set. i,e., $D_{C} \subset D_{T}$.}, while~\cite{gondal2021function, mathieu2021contrastive} validate the model capacity in predicting the parameters of data-generating functions $f$ using only $C$ {\it context} samples.

\subsection{Energy-based Model}
Energy-based models~(EBMs)~\cite{lecun2006tutorial} are a group of density estimators with {\it implicit} density functions.
Denote $\mbf{y}$ as any i.i.d observations sampled from $\bs{y}_{1:T}$.
Let $\varphi: \mbb{R}^{d_{\mca{Y}}} \to \mbb{R}$ be the energy function parameterized by an arbitrary neural network.
EBMs formulate the density of $\mbf{y}$ as a Boltzmann distribution,
\begin{equation}\label{eq: ebm}
    p(\mbf{y}; \varphi) = \frac{ \exp\left\{ - \varphi(\mbf{y}) \right\} }{ Z(\varphi) },
\end{equation}
where $Z(\varphi) = \int \exp\left\{ - \varphi(\mbf{y}) \right\} \dev \mbf{y}$ is the partition function with respect to $\mbf{y}$, allowing $\varphi(\mbf{y})$ to output unnormalized scalar. 
This flexibility in designing a neural energy function translates into the expressivity of EBMs in modeling high-dimensional data~\cite{du2019implicit} but also precludes direct samplings from $p(\mbf{y}; \varphi)$, typically requiring expensive and biased MCMC~\cite{tieleman2008training, nijkamp2019learning}.

As an alternative, Noise contrastive estimation~(NCE)~\cite{gutmann2010noise} has widely been adapted to learn unnormalized models like EBMs~\cite{gao2020flow, aneja2021contrastive}.
In principle, NCE fits EBMs to an unknown ground-truth distribution by learning to distinguish true samples from noise samples generated from a reference noise distribution.
Observe $N$ samples $\bs{y}_{1:N}$ collected from a mixture of data $p_{d}(\mbf{y})$ and noise distribution $p_{n}(\mbf{y})$ defined with a Bernoulli indicator: $p_{d, n}(\mbf{y}) = p(v=0) p_{n}(\mbf{y}) + p(v=1) p_{d}(\mbf{y})$.
Consider an EBM, which absorbs $Z(\varphi)$ as the model's trainable parameters, now has a normalized likelihood $p(\mbf{y} ; \varphi)$.
In this case, EBMs estimate the true data distribution $p_d(\mbf{y})$ by training a classifier $\varphi(\mbf{y})$
that minimizes the binary cross-entropy objective,
\begin{equation}\label{eq:nce}
    \min_{\varphi} - \left[ \mathbb{E}_{\mbf{y} \sim p_{d}(\mbf{y})} \log \varphi(\mbf{y}) + \mathbb{E}_{\mbf{y} \sim p_{n}(\mbf{y})} \log \left( 1 - \varphi(\mbf{y}) \right)  \right].
\end{equation}
Essentially, EBM reaches optimality if its predictive posterior matches the true posterior $\varphi^{*}(\mbf{y}):= p(v=1 | \mbf{y}; \varphi) = p_{d,n}(v=1 | \mbf{y})$, i.e., correctly estimates the noise-data ratio of the mixture.
By doing so, density estimation for EBMs can be viewed as a classification problem, assuming a noise distribution with an explicit and tractable density function.

\section{Method}
\label{sec:method}

Our method involves implementing a joint training approach to combine CNPs with EBMs in light of their complementary attributes under the NCE framework.
CNPs secure tractable and exact density estimation by locking up an explicit likelihood function by definition.
In exchange, this comes at the cost of~(possibly) inappropriate assumptions for observations $\mbf{y}$ with high dimensions~\cite{kim2021multi}.
In comparison, EBMs can be flexibly formulated in terms of network structure $\varphi$ to encompass complex distributions; yet, requiring a strong reference distribution upon performing NCE.
Adversarial training~\cite{goodfellow2020generative} provides an effective bridge to connect CNPs with EBMs in the NCE context, where CNP resembles the generator while EBM acts as a critic. 
CNP receives feedback on their generations from a discriminative perspective, while EBM also benefits from stronger predictions given by improved CNP.
We now discuss a two-stage adversarial training scheme.

\begin{figure}
    \begin{center}
    \includegraphics[width=0.9\linewidth]{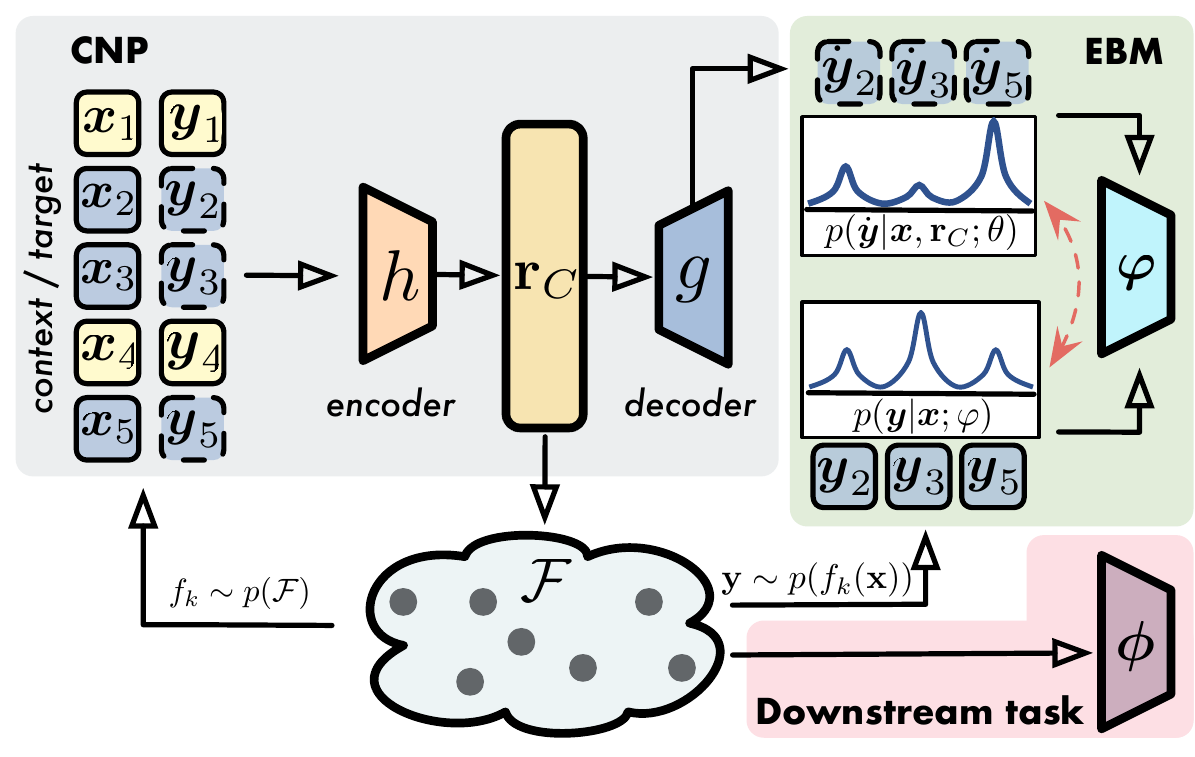}
    \caption{Method illustration. Stage-1) regular training of CNP $\theta$. Stage-2) calibrate CNP $\theta$ with EBM $\varphi$. 3) after training, extract {\it context} representation $\mbf{r}_{C}$ from frozen CNP to finetune the prediction head $\phi$ for downstream tasks.}
    \label{fig:1_framework}
    \end{center}
    \vskip -0.2in
\end{figure}

\subsection{Training CNPs as a Noise Distribution}
\label{subsec:train_cnp}
The effectiveness of NCE depends on how close the noise distribution is to the ground truth~\cite{gutmann2010noise, gao2020flow, aneja2021contrastive}:
if $p_{n}(\mbf{y})$ is too far from $p_{d}(\mbf{y})$, EBM cannot obtain much useful information since the classifier $\varphi(\mbf{y})$ can easily distinguish true observations from noises.
It is thus crucial to train CNP to be a strong noise distribution before NCE training.
For this stage, we follow the regular training of CNPs shown in~\cref{fig:1_framework}.
Consider each function instantiation $f_k \sim F_{train}$, CNPs collapse\footnote{\cref{eq:cnp_fac} does not necessarily imply tractability, which however is determined by the specific design choice of $p(f | D_C ; \theta)$} the empirical prior $p(f_k | D_C^{k}, \theta)$ into a deterministic {\it context} representation $\mbf{r}_{C}^{k}$, approximated by a neural network-based {\it encoder} $h$ applied to the {\it context} set,
\begin{equation}
    \mbf{r}_{C}^{k} = \oplus_{c=1}^{C} \bs{r}_{c}^{k}, \; \text{with } \bs{r}_{c}^{k} = h \left( \bs{x}_{c}^{k}, \bs{y}_{c}^{k} \right)
\end{equation}
where $\oplus$ is a permutation-invariant aggregate to satisfy exchangeability and consistency.
Typical implementations include mean-pooling by CNP~\cite{garnelo2018conditional} and self-attention by AttnCNP~\cite{kimanp18}.
Following, another parameterization in CNPs involves a {\it decoder} $g$ that computes sufficient statistics of the predictive distribution of each {\it target} observations $\bs{y}_{t}^{k}$, conditioned on $\bs{x}_{t}^{k}$ and $\mbf{r}_{C}^{k}$.
That is, $g$ parameterizes the mean $\bs{\mu}_t$ and variance $\bs{\Sigma}_t$ for the most heavily used Gaussian.
In this sense, the parameters of CNPs $\theta = \{ h, g \}$ can be optimized by MLE with a factorized exact log-likelihood over $ \{ \bs{y}_{1:T}^{k} \}_{k=1:K} $,
\begin{equation}\label{eq:cnp}
    \begin{aligned}
        \max_{\theta} \; & \mbb{E}_{f_k \sim F_{train}} \left[ \log p(\bs{y}_{1:T}^{k}, \bs{x}_{1:T}^{k} \, ; \, \theta)  \right] \\
                         & = \frac{1}{K} \sum_{k=1}^{K} \left\{ \sum_{t=1}^{T} \log \mca{N}(\bs{y}_{t}^{k} | \bs{x}_{t}^{k}, \mbf{r}_{C}^{k} \, ; \bs{\mu}_{t}^{k}, \bs{\Sigma}_{t}^{k}) \right\} \, ,
    \end{aligned}
\end{equation}
with $(\bs{\mu}_{t}, \bs{\Sigma}_{t}) = g (\bs{x}_t, \mbf{r}_{C})$ and $ \bs{y}_{t}^{k} \indep \bs{y}_{t^{\prime}}^{k} \, | \, \mbf{r}_{C} \; \text{for } k \neq k^{\prime} $.
One can set the likelihood to other distributions depending on the dataset, e.g., Bernoulli as in~\cite{norcliffe2021neural}.

Upon achieving a steady low value of log probability, we consider CNP strong enough to be a noise generator.
Now we begin calibrating CNP with EBM.

\subsection{Calibrating CNP with EBM}
Recall that our purpose is to lift the CNP's expressivity with a flexible EBM.
We do not follow~\cite{aneja2021contrastive} where the parametric noise distribution is fixed after stage-1 training, rather in a similar way to~\cite{gao2020flow} that updates the CNP and EBM parameters in an adversarial manner.
One can relate the motivation here to the training in Generative Adversarial Networks: 
provided a powerful EBM, such as one that can estimate the density ratio correctly, means that CNP does not generate ``realistic'' reconstructions close enough to the real observations, and would need to be updated in order to confuse the EBM.
On the other hand, CNP would only gain little cost for model improvement in the event that EBM labels the true observations as fake.
As such, we can adapt the NCE objective to the joint optimization of EBM and CNP, in the way of playing a minimax game until the optimization reaches Nash equilibrium.
The CNP predictions can be thought optimal if EBM can no longer identify them from true function observations.

Following a CNP forward pass with function instantiations $\{ f_{k} \}_{k=1:K}$,
we have $K * T$ generated observations $\dot{\bs{y}}_{1:T}^{k}$ by CNP, and $K * T$ ground-truth observations $\bs{y}_{1:T}^{k}$ over all the $K$ instantiations.
We thus obtain a mixture of true data and noises to be learned from as in~\cref{eq:nce}.
This allows us to set a noise-to-data ratio ranging from $1$ to $K$.

We let EBM and CNP iteratively maximize and minimize the posterior log-likelihood of an empirically approximated NCE objective as,
\begin{equation}\label{eq:cnp_nce}
    \begin{aligned}
        & \min_{\theta} \max_{\varphi}  \\
        & \sum_{k=1}^{K} \left\{ \sum_{t=1}^{T} \log \frac{ p(\bs{y}_{t}^{k} | \bs{x}_{t}^{k}; \varphi)  }{p(\bs{y}_{t}^{k} | \bs{x}_{t}^{k}; \varphi) + p(\bs{y}_{t}^{k} | \bs{x}_{t}^{k}; \theta)} \right. + \\
        & \left. \sum_{j=1}^{K} \sum_{t=1}^{T} \log \frac{ p(\dot{\bs{y}}_{t}^{j} | \bs{x}_{t}^{j}; \theta)  }{p(\dot{\bs{y}}_{t}^{j} | \bs{x}_{t}^{j}; \varphi) + p(\dot{\bs{y}}_{t}^{j} | \bs{x}_{t}^{j}; \theta)} \right\} - \log K
    \end{aligned}
\end{equation}
where $p(\bs{y}_{t}^{(k)} | \bs{x}_{t}^{(k)}; \varphi)$ and $p(\dot{\bs{y}}_{t}^{(k)} | \bs{x}_{t}^{(k)}; \varphi)$ are estimated log-probabilities of true and CNP's predictive observation at covariate $\bs{x}_{t}$ of the $k$-th function, calculated by EBM.
Similarly, CNP computes log-probabilities $p(\bs{y}_{t}^{(k)} | \bs{x}_{t}^{(k)}; \theta)$ and $p(\dot{\bs{y}}_{t}^{(k)} | \bs{x}_{t}^{(k)}; \theta)$ of true and predicted observations accordingly.
As for iterative training in practice, we set a threshold $\alpha$ to determine either EBM or CNP to be optimized, based on the accuracy of the density ratio estimated by EBM.
We fix EBM and solely optimize for CNP in the case of better accuracy than $\alpha$, and vice versa.

\subsection{Facilitating NCE with CNP}
CNP is not the only beneficiary of adversarial training.
The optimality of noise can reversely affect the performance of NCE estimator~\cite{chehab2022optimal}.
All throughout training, CNP provides the MLE estimate of noise parameters thanks to the exact density likelihood.
Using the estimate of noise distribution is shown to be helpful in reducing the asymptotic variance of NCE~\cite{uehara2018analysis}.

Meanwhile, the large batch training required by previous contrastive learning models~\cite{oord2018representation, chen2020simple} may also be appreciated in this setting.
Note that the summation from $1$ to $K$ with the second term inside brackets in~\cref{eq:cnp_nce} means that we can get negative samples from the CNP's predictions of other $K - 1$ functions.
While NCE could theoretically reach Cramér-Rao bound with an infinitely large batch size~\cite{gutmann2012noise}, it is computationally infeasible.
However, one can also set this noise-to-data ratio to $1$ following common practices, i.e., some proportion for true observations and noises.
With the predictions of CNP, this can be touched by only contrasting the ground-truth to predictions {\it within} the same instantiation.

\subsection{Regularizing MLE with NCE}
Following the results of~\cite{goodfellow2020generative}, prior works~\cite{gao2020flow, aneja2021contrastive} connect the NCE objective in~\cref{eq:cnp_nce} to an approximate minimization of Jensen-Shannon~(JS) divergence between true distribution of observations $p(f(\mbf{x}))$ and predictive density of CNP $p(\mbf{y} | \mbf{x}; \theta)$ when updating $\theta$,
\begin{equation}
    D_{JS} \left( p(\mbf{y} | \mbf{x}; \theta) \, || \, p(f(\mbf{x})) \right) = \left[ \varphi^{*}(\mbf{y}) - \log (4) \right] / \, 2 \, ,
\end{equation}
provided that EBM is the optimal $\varphi^{*}$, i.e., the density of EBM equals to ground-truth $p(\mbf{y} | \mbf{x}, \varphi^{*}) = p( f(\mbf{x}))$.
This may suggest that if the minimax game is successfully played, NCE could facilitate the estimation of CNP other than simply optimizing the KL-divergence as in MLE objective.

Yet, we find that the performance of CNP is empirically hindered if we fully turn to~\cref{eq:cnp_nce}~(\cref{appendix:obj_ratio}).
Thus, we derive a more practical objective in implementations by weighting~\cref{eq:cnp} with~\cref{eq:cnp_nce}.
Denote the equivalent loss of~\cref{eq:cnp} to $\ell_{1}(\theta)$ and~\cref{eq:cnp_nce} to $\ell_{2}(\varphi, \theta)$, we conclude the loss function in stage-2 as $(1 - \beta) \cdot \ell_{1}(\theta) + \beta \cdot \ell_{2}(\varphi, \theta)$, where $\beta$ is the trade-off coefficient.
To this extent, we can view NCE as an auxiliary objective of MLE optimization of CNPs.

\subsection{Downstream Tasks Transfer}
While the post-training capability of {\it context} representations has not been extensively evaluated in the CNPs literature, this appears to be another interest metric as discussed in~\cite{gondal2021function, mathieu2021contrastive}.
In this paper, we consider the supervised finetuning evaluations on the {\it meta-test} set $F_{test}$.
Assume each sample in $F_{test}$ is of the form $(f_{e}, l_{e})$.
Each instantiation has an associated label.
For example, in the case of Motor Imagery recognition with electroencephalography~(EEG) signals.
Let $f_{e}$ be a recorded EEG clip recorded when a subject is performing a particular imagination class $l_{e}$~(e.g., imaging making a fist with the left hand).
Using a pre-trained CNP with frozen parameters $\theta^{*}$, we extract the {\it context} representation $\mbf{r}_{C}$ from the {\it context} set, and input it to the prediction head $\phi( \mbf{r}_{C}): \mbb{R}^{d_{r}} \to \mbb{R}^{d_{l}}$.
The prediction head follows standard supervised optimization.
In this sense, training CNPs can be seen as self-supervised representation learning.
Code is available at~\cref{link:code}.

\section{Related Work}
\label{sec: literature}


\textbf{Conditional Neural Processes~(CNPs)}.
CNPs were proposed to bridge the expressivity of neural networks and the fast adaptability of non-parametric models, defining their inference as encoding {\it context} samples, aggregation, and decoding over {\it target} samples with a factorized predictive distribution.
As CNP~\cite{garnelo2018conditional} does not consider any correlations between samples, the subsequent works aim to improve model expressivity.
The CNPs members contribute to leveraging distances between samples.
AttnCNP~\cite{kimanp18} uses self-attention within {\it context} set and cross-attention with respect to each {\it target} point to account for absolute distances between covariates.
Thereafter, ConvCNP~\cite{Gordon2019ConvCNP} and SteerableCNP~\cite{holderrieth2021equivariant} introduce symmetries into CNPs, namely translation equivariance and group equivariance, respectively.
Another family relating to CNPs has evolved since~\cite{garnelo2018neural}, latent NPs members replace (or append to) the deterministic inference of CNPs with latent variables to capture the global uncertainty of data-generating processes and transform the predictive distribution into an infinite mixture of Gaussian.
Further works draw inspiration from hierarchical generative models and equip NPs with increasingly powerful latent hierarchies~\cite{kim2021multi, wang2022learning}.
Yet, improved capacity also comes with a cost of intractable likelihood.
One needs to apply a variational lower bound~\cite{garnelo2018neural, kimanp18} or repeat Monte Carlo approximations~\cite{Foong2020ConvNP} to derive the approximated log-likelihood.
Recently, TNP~\cite{nguyen2022transformer} applies a casually-masked transformer to accommodate sequential data, introducing autoregressive predictions to NPs.
Instead of presenting a {\it new variant} of CNPs, our work discusses a {\it general calibration} principle with adversarial training, which is compatible with any CNP member.
Our method neither alters CNP's original inference procedure nor burdens it with heavy computation overhead.

\textbf{Function Contrastive Representation Learning~(FCRL)}.
CNPs had previously combined noise contrast estimation (NCE) in the spirit of contrastive self-supervised learning.
Inspired by MetaCDE~\cite{ton2021noise}, FCRL~\cite{gondal2021function} pursues a robust {\it meta}-representation space for {\it context} set, where partial views of the same function are pulled closer while different functions are pushed away.
CReSP~\cite{mathieu2021contrastive} replaces the mean-pooling of {\it context} representation with self-attention.
While FCRL and CReSP entirely discard generation and prioritize downstream tasks where reconstruction is unnecessary, NCE optimization may be used as regularization to the generation~\cite{kallidromitis2021contrastive, ye2022contrastive}.
Still, these practices do not address the question, as to how {\it context} representations can be improved to facilitate generation.
Despite also using NCE, our motivation differs from all these works as we {\it directly} calibrate the CNPs' generation.
We consider the predictive distribution to be noise in NCE and refine CNP with a more flexible unnormalized estimator.

\textbf{Energy-based Models~(EBMs)}.
EBMs are arguably flexible statistical models in terms of their unconstrained function form and unnormalized output.
The expressive power of EBMs is demonstrated in recent connections to deep models~\cite{du2019implicit}.
A compositional nature is also evident when EBMs are combined with explicit generative models such as variational autoencoders~(VAEs)~\cite{kingma2013auto}.
VAEs can be taught a prior distribution with EBMs empowered by implicit generation~\cite{pang2020learning, xiao2020vaebm, aneja2021contrastive}.
Yet, the downside of these approaches involves expensive MCMC sampling with high-dimensional data.
To avoid implicit sampling, NCE is widely studied for estimating EBMs by separating ground-truth data density from some sample-efficient reference distributions referred to as noise.
It is primarily important to find an appropriate noise distribution.
Conditional NCE~\cite{ma2018noise} generates noises with observed true data.
Bose et al. propose an adversarially learned sampler from the mixture of true data and noise~\cite{bose2018adversarial}.
Our work takes a similar spirit to~\cite{gao2020flow}, where EBMs are applied to help the estimation of a normalization flow of true data.
In contrast, our work is devoted to learning a prediction map in the meta-learning setting, where noise samples can be obtained via a single forward pass of CNPs.

\section{Empirical Studies}
\label{sec: experiments}

We plug the proposed method into a range of CNP variants to justify the performance change before and after adversarial calibration.
We evaluate the model capacity in terms of generation performances~(as in most CNPs literature;~\cref{subsec:generation}) and the transferability of {\it context} representations to four downstream tasks~(\cref{subsec:downstream}).
Both evaluations involve synthesized and real-world data.
Our interests fall into two central questions: (1) Does adversarial training with noise contrastive estimation improve the generation performance of CNPs? (2) If so, does improved generation ability promote downstream tasks such as classification?

\subsection{Generation}\label{subsec:generation}
\subsubsection{Synthesized 1D Data Regression}\label{subsubsec:1d_gen}
\begin{table}[t]
\caption{The regression MSE ($\downarrow$) on synthetic 1D GP functions, sine waves, and damped oscillators$(\times 10^{-1})$. {\bf Bold} indicates adversarial training leads to performance improvement over the original CNP model. Results are mean $\pm$ std with three different seeds. }
\label{tab:result_gp}
\begin{center}
\begin{small}
\resizebox{\columnwidth}{!}{%
\begin{tabular}{lcccc}
\toprule
\textbf{Method}                  & \textbf{GP-RBF}  & \textbf{Sinusoid}  & \textbf{Oscillator} \\
\midrule
CNP                     & 2,873 $\pm$	0.162  & 0.795 $\pm$ 0.151      &  0.246	$\pm$ 0.029              \\
ACNP                    & 1.829	$\pm$ 0.117  & 1.138 $\pm$ 0.062     &   0.697 $\pm$ 0.067             \\
CCNP                    & 2.185	$\pm$ 0.072   & 0.459 $\pm$ 0.026     &  0.596	$\pm$ 0.133              \\
\midrule
CNP-adv                   & \textbf{2.107}	$\pm$ 0.096   & \textbf{0.481} $\pm$ 0.042      & \textbf{0.183}	$\pm$ 0.025               \\
ACNP-adv                  & \textbf{1.507}	$\pm$ 0.192   & \textbf{0.987} $\pm$ 0.073      & 0.711 $\pm$ 0.081               \\
CCNP-adv                  & \textbf{2.093}	$\pm$ 0.015   & \textbf{0.389} $\pm$ 0.042      & \textbf{0.221} $\pm$	0.022               \\
\bottomrule
\end{tabular}}
\end{small}
\end{center}
\vskip -0.2in
\end{table}

\textbf{Setup}.
We first consider a few-shot regression with 1D synthetic data generated by Gaussian Processes~(GPs) with RBF kernels, and 1D functions generated by closed-form expressions of sine waves and damped oscillators.
For GP-RBF data, we randomly generate 4,096 functions for {\it meta-train}, and 1,000 functions for {\it meta-test}.
Each function instantiation has 128 points.
In training, the {\it context} set is randomly sampled from $\mca{U}[0.1, 0.3]$, all 128 points are used as the {\it target} set, while in evaluation, the {\it context} size is fixed to 0.3.
As for sine waves and oscillators, we set the {\it meta-train} functions and {\it meta-test} functions to 500, where each instantiation has 100 points.
As well, the {\it context} size is between $[0.04, 0.2]$ during training and is fixed to $0.2$ in evaluation.
For sine waves, we vary the values of amplitude $\mca{U}[-1, 1]$ and shift within $\mca{U}[-0.5, 0.5]$, covariates $x \in [-3\pi, 3\pi]$.
Each oscillator function has a unique amplitude and shift $a \sim \mca{U}[-3, 3]$ and $b \sim \mca{U}[-1, 1]$, with $x$ starting at $0$ and going up to $5$.
See full details of the setup in~\cref{appendix:sine_oscil_exp}.

\textbf{Results}.
\cref{tab:result_gp} presents the quantitative results of various baseline CNPs with and without adversarial training.
The baseline models include CNP~\cite{garnelo2018conditional}, ACNP~\cite{kimanp18}, and CCNP~\cite{kallidromitis2021contrastive}, where the first two are commonly adopted as per most CNPs literature.
We consider CCNP as a comparison to the previous NCE setting, where NCE is also applied but on the {\it context} set as a regularization term to CNPs.

We find that adversarial training consistently results in performance improvements to baselines in most cases.
Notably, while ACNP is arguably more powerful than CNP and indeed reports better results as expected for GP data~\cite{kimanp18}, it performs worse with sine and oscillator data, probably due to the number of functions being too limited.
Also, CCNP does not improve CNP consistently either.
Further, we observe that adversarial training can be adapted to CCNP, showing potential in improving generative CNPs by leveraging composite contrastive objectives.

\subsubsection{Basketball Trajectory Imputation}


\begin{table}[t]
\caption{The test likelihoods comparison ($\uparrow$) with basketball game trajectories, without and with adversarial training. Results are with three different seeds, improved results are in {\bf bold}.}
\label{tab:result_basket}
\vskip -0.5in
\begin{center}
\begin{small}
\resizebox{\columnwidth}{!}{%
\begin{tabular}{lccc}
\toprule
                  & \textbf{CNP}  & \textbf{CCNP}  & \textbf{ACNP} \\
\midrule
w/o adv                     & -26.266 $\pm$	8.126  & -3.457 $\pm$ 2.456      &  216.268	$\pm$ 0.5026              \\
w/ adv                    & \textbf{-0.947}	$\pm$ 1.961  & \textbf{17.304} $\pm$ 4.277     &   \textbf{221.016} $\pm$ 2.771             \\
\bottomrule
\end{tabular}
}
\end{small}
\end{center}
\vskip -0.2in
\end{table}

\textbf{Setup}.
We next investigate a real-world trajectory dataset, the STATS SportsVU tracking
data from the 2012-2013 NBA season~\cite{lucey2014get}.
Each sequence describes the movements of 10 players and the ball with respect to their coordinates on the court, lasting for 50 timesteps.
We are tasked with imputing the observation of the coordinates at each timestep, where each observation is of dimension $\mbf{y} \in \mbb{R}^{22}$.
The {\it meta-train} set and {\it meta-test} contains 104,003 and other 13,464 sequences, respectively.
For each batch during training, the {\it context} size is randomly sampled $\mca{U}[0.1, 0.5]$ whilst using all 50 frames as the {\it target} set.
The {\it context} size in evaluation is fixed to $0.5$.
We predict the outcome with diagonal multivariate normal distribution
because in basketball games all players take action in relation to the ball.
More details are covered in~\cref{appendix:basket}.

\textbf{Results}.
Observe that adversarial training consistently impacts all CNP variants positively from~\cref{tab:result_basket}, especially for CNP and CCNP.
By introducing adversarial training, the predictive likelihoods of CNP and CCNP improve by orders of magnitude over the original model, from $-26.27$ to $-0.95$ and $-3.46$ to $17.30$, respectively.
CCNP regularizes CNP with NCE by pairing different partial views of each function, we find that adversarial training further improves the results of CCNP.
ACNP significantly outperforms the other two CNP variants with and without adversarial training thanks to the attention mechanism.
Nevertheless, it is possible for ACNP to be overconfident in its predictions~\cite{nguyen2022transformer}.
Such cases may not benefit ACNP as much as CNP and CCNP from adversarial calibration.

\subsubsection{Image Sequence Reconstruction}

\begin{figure}[t]
    \begin{center}
    \centerline{\includegraphics[width=0.9\columnwidth]{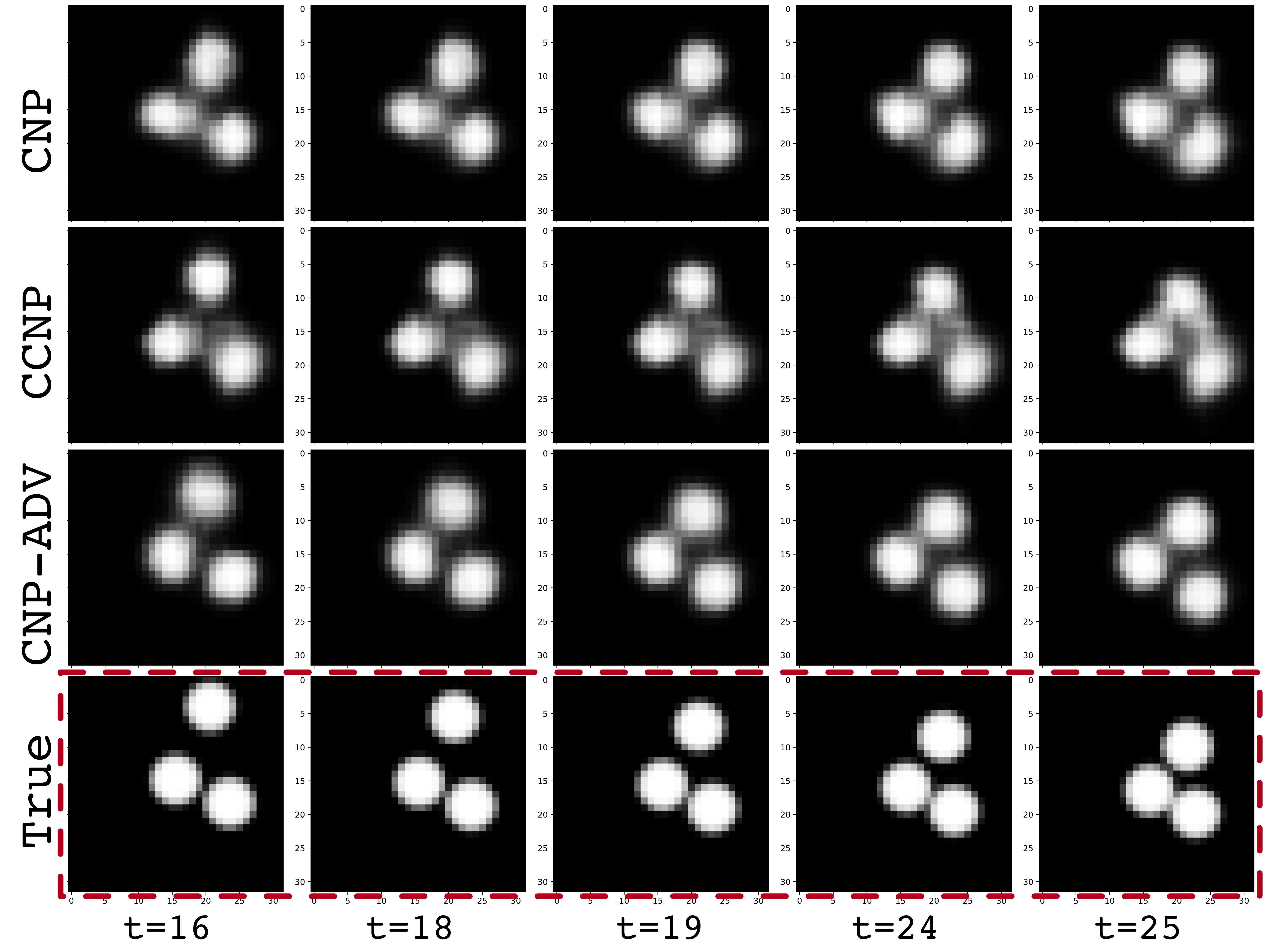}}
    \caption{The qualitative comparison with bouncing ball data. The showcased frames are {\it irregularly} sampled within one sequence.}
    \label{fig:bball}
    \end{center}
    \vskip -0.4in
\end{figure}

\textbf{Setup}.
The third generation task involves reconstructing irregularly sampled high-dimensional sequences of images~\cite{yildiz2019ode2vae}.
The dataset describes a bouncing ball system with three interacting balls moving within a rectangular box.
Each sequence contains 50 frames of $32\times32$ grayscale images.
The {\it meta-train} set has 10,000 sequences while we use another 500 sequences in {\it meta-test} set.
We randomly sample the {\it context} size from $\mca{U}[0.1, 0.5]$ during training and fix that to 0.5 for evaluation.
Following~\cite{norcliffe2021neural}, we set the predictive likelihood to diagonal Bernoulli since most of the pixel intensities are located around $0$ and $1$.
We include the full details in~\cref{appendix:bball}.

\textbf{Results}.
\cref{fig:bball} qualitatively compares the generation results of CNP, CCNP~(i.e., with NCE regularization), and CNP-adv~(i.e., with adversarial training).
As seen from the first and the second rows, CNP and CCNP cannot cope with frames where three balls are gathered closely.
The reconstructions are blurry, and even balls cannot maintain the shape in, e.g., $t=24, 25$.
In contrast, CNP-adv generates better-quality images.
Meanwhile, we directly evaluate the effects of adopting NCE with adversarial training or as regularization to the {\it context} set.
We thus conclude that NCE combined with adversarial calibration improves the capability of CNP in reconstructing high-dimensional observations.

\subsection{Downstream Tasks}\label{subsec:downstream}
In this section, we evaluate if NCE indeed helps with the transferability of {\it context} representations.
The pre-trained CNPs are used as self-supervised feature extractors~(parameters are frozen), followed by a trainable prediction head $\phi(\cdot)$ upon performing supervised downstream prediction tasks.
\subsubsection{Function Parameter Inference}


\begin{figure}[t]
    \subfigure[Oscillator amplitude $a$ \label{fig:fn_params_a}]{
        \includegraphics[width=0.47\columnwidth]{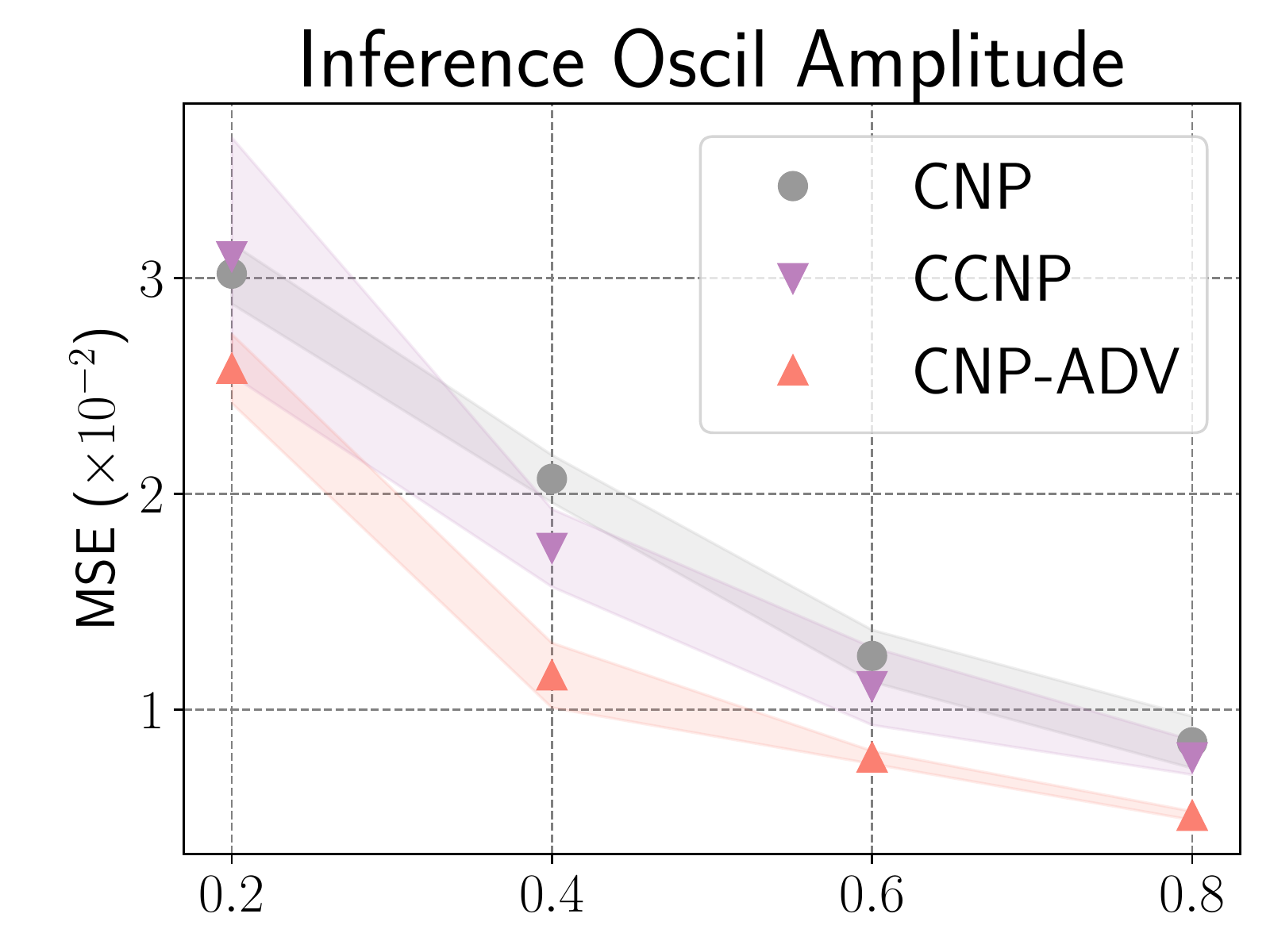}
    }
    \subfigure[Oscillator shift $b$ \label{fig:fn_params_b}]{
        \includegraphics[width=0.47\columnwidth]{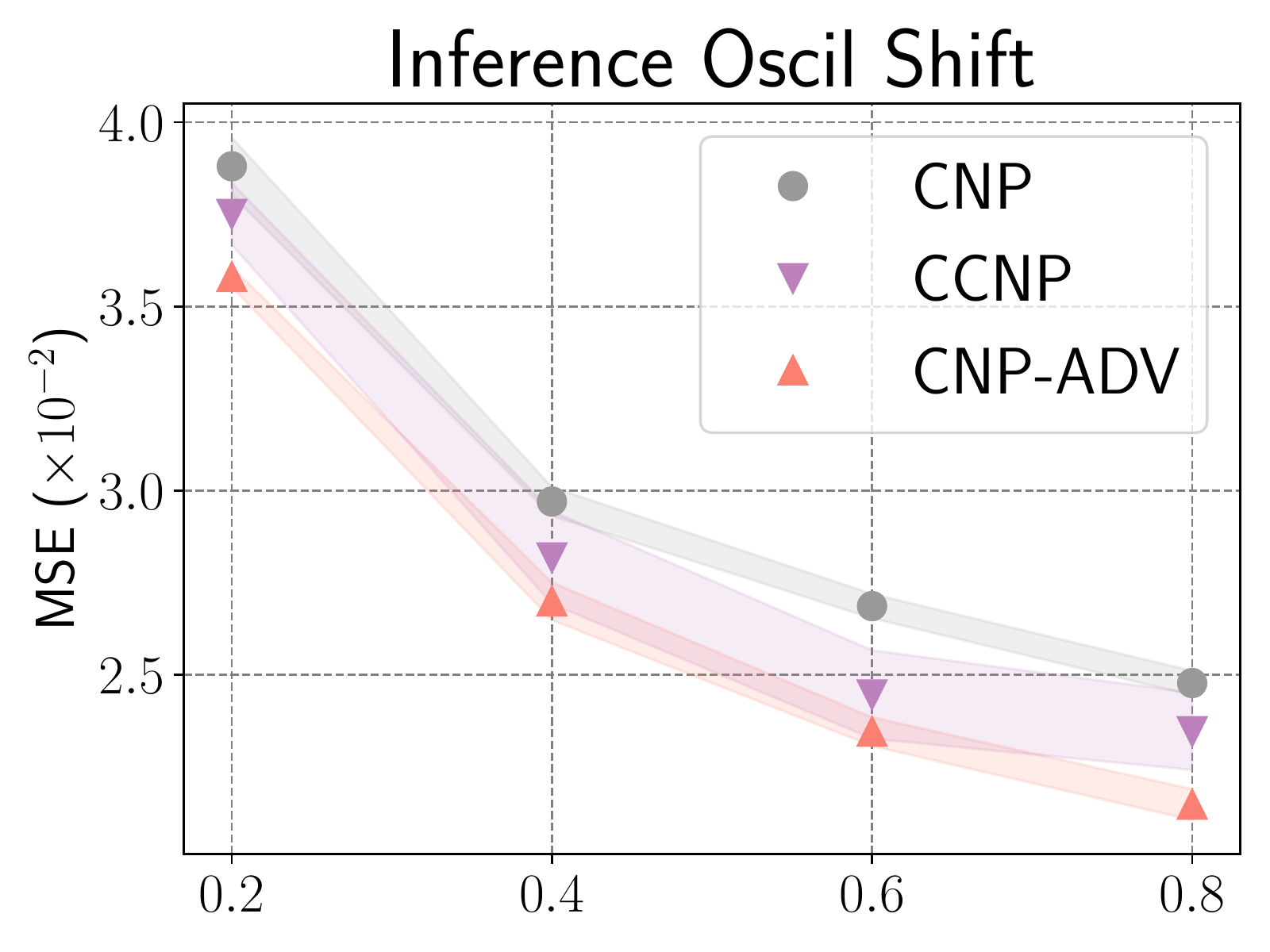}
    }
    \caption{The prediction MSE~(mean$\pm$std) $\downarrow$ of 1D function parameter inference on Oscillator data, with different {\it context} sizes.}
    \label{fig:fn_params}
    \vskip -0.2in
\end{figure}

\textbf{Setup}.
For the CNPs trained with the oscillator dataset in~\cref{subsubsec:1d_gen}, we append a prediction head to infer the amplitude $a$ and shift $b$.
Given some {\it context} points, the task is to identify a specific $a$ and $b$ used to generate the function.
The downstream predictions sample random {\it context} points in a fixed number for both training and testing.
The best-performing CNPs of each run are used.
The prediction head is trained for 20 epochs.
\cref{appendix:sine_oscil_exp} covers a detailed setup.

\textbf{Results}.
While CNPs are pre-trained with the {\it context} size $\mca{U}[0.04, 0.2]$, they can handle varied numbers of {\it context} points in evaluation.
\cref{fig:fn_params} depicts the regression MSE with different {\it context} size ranging from $0.2$ to $0.8$.
More context points lead to improved prediction performances for all the CNP, CCNP, and CNP-ADV.
Both CCNP and CNP-ADV include NCE, shown to help CNP identify the underlying function parameter.
Although this is a typical use case of improving the {\it context} set with NCE~\cite{gondal2021function}, we observe that CNP-adv consistently outperforms CCNP, even without focusing on the {\it context} set particularly.

\subsubsection{Motor Imagery Classification}

People-centric sensing can be an important real-world application of CNPs, where the recordings are naturally noisy and contain missing values frequently.

\textbf{Setup}.
We predict the associated motor imagery task based on irregularly sampled Electroencephalography~(EEG) measurements~\cite{goldberger2000physiobank}.
The dataset contains 64-channeled EEG recordings of 105 subjects performing various motor imagery tasks.
We perform {\it binary classification} regarding if opening and closing both {\it fists} or {\it feet}.
EEG clips are 4 seconds long with 640 timesteps.
The {\it context} size is sampled from $\mca{U}[0.5, 0.8]$ during CNP training.
We split all the clips of 105 subjects with a ratio of 3,746/400/400 clips in train/validation/test sets.
The prediction head is trained for 10 epochs.
See~\cref{appendix:eeg} for more details.

\begin{table}[t]
    \caption{The prediction accuracy with EEG Motor Imagery dataset.}
    \begin{center}
        \begin{small}
        \begin{tabular}{cccc}
        \toprule
             & CNP & CCNP & CNP-ADV \\
        \midrule
         MSE($\downarrow$) & 1.99 $\pm$ 0.24     &  1.78 $\pm$ 0.44    & \textbf{1.68} $\pm$ 0.16       \\
         ACC($\uparrow$) & 66.46 $\pm$ 1.40    &  65.18 $\pm$ 1.90    &  \textbf{70.43} $\pm$ 0.78      \\
         \bottomrule
        \end{tabular}
        \end{small}
    \end{center}
    \label{tab:my_label}
    \vskip -0.2in
\end{table}

\textbf{Results}.
Similarly, both CCNP and CNP-ADV contribute to more accurate reconstruction of EEG measurements.
However, CCNP appears to perform worse in the Motor Imagery classification task.
In this case, the increased reconstruction ability does not necessarily lead to a {\it context} representation that facilitates downstream tasks, which is aligned with~\cite{gondal2021function}.
While CCNP draws the motivation therefrom, the NCE adopted on {\it context} set does not clearly demonstrate effectiveness in practice.
A possible reason is, the number of functions~(3,746) might be insufficient to express the capacity of self-supervised contrastive learning.
In contrast, CNP-ADV decreases the reconstruction MSE while also increasing the classification accuracy.
Accordingly, the adversarial calibration of CNPs with NCE could benefit both reconstruction and downstream tasks.

\subsubsection{Human Activity Recognition}

\textbf{Setup}.
We lastly evaluate the activity recognition capacity with PAMAP2~\cite{reiss2012introducing}, another sensor-based people-centric dataset that contains records of 8 subjects upon performing 4 different physical activities, with 3 inertial measurement units~(IMU).
Each IMU records 12 effective features per timestep.
The activity labels contain 
Following mild preprocessing, e.g., drop invalid values and apply sliding windows, we obtain 50 timesteps times with $\mbf{y} \in \mbb{R}^{36}$ for each sequence.
The {\it context} size subjects to $\mca{U}[0.5, 0.7]$ during training.
For each run, we select 2 subjects for validation and testing.
The remaining 6 subjects are used for pre-training CNPs and then training the prediction head.
This results in around 9,600 training, 1,600 validation and 1,600 testing sequences.
The prediction head is trained for 10 epochs.
\cref{appendix:pamap} covers more details of the setup.

\begin{figure}[t]
    \subfigure[Average Acc of 8 subjects \label{fig:pamap_overall}]{
        \includegraphics[width=0.47\columnwidth]{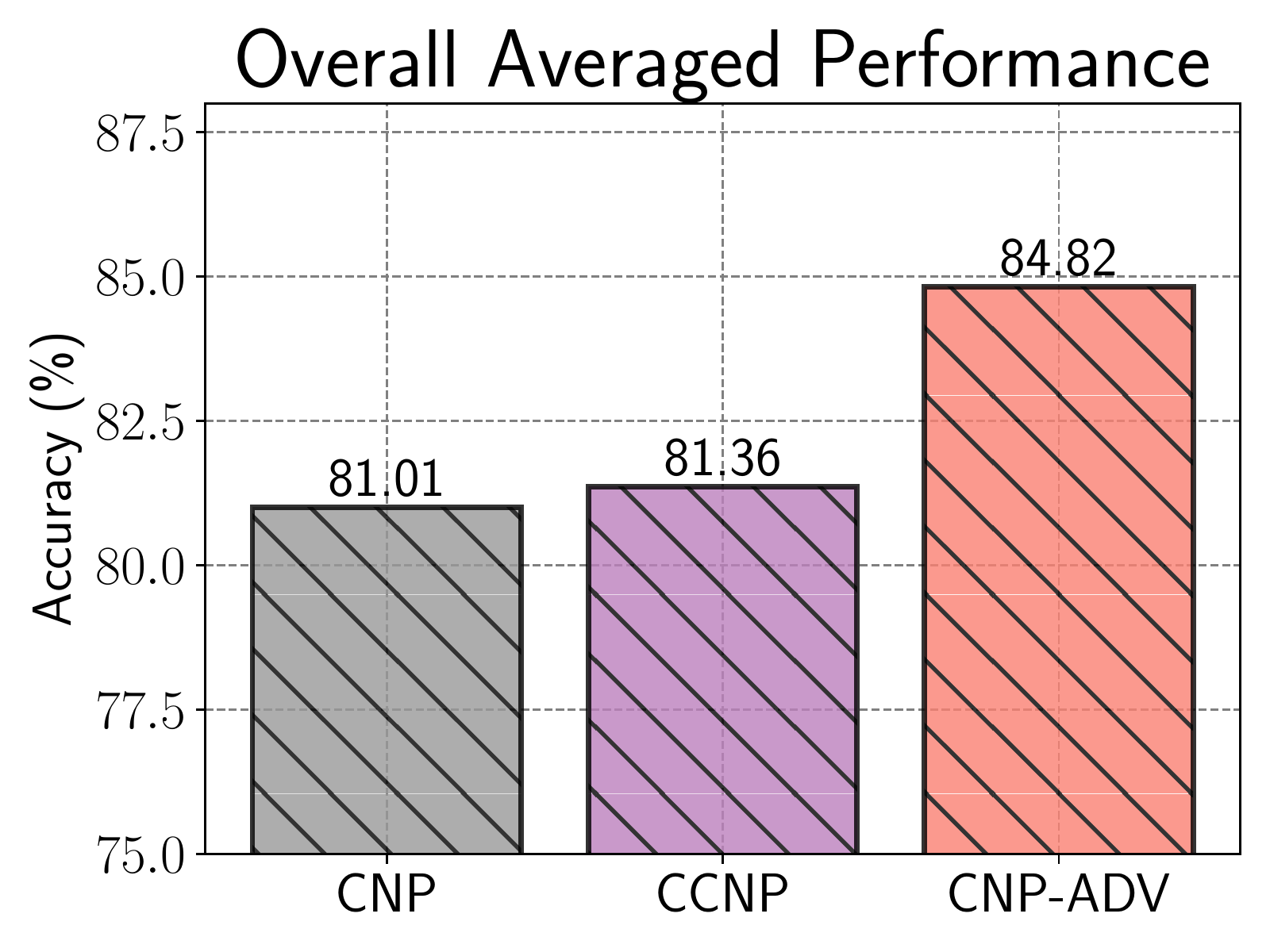}
    }
    \subfigure[Subject-specific Acc \label{fig:pamap_sensitivity}]{
        \includegraphics[width=0.47\columnwidth]{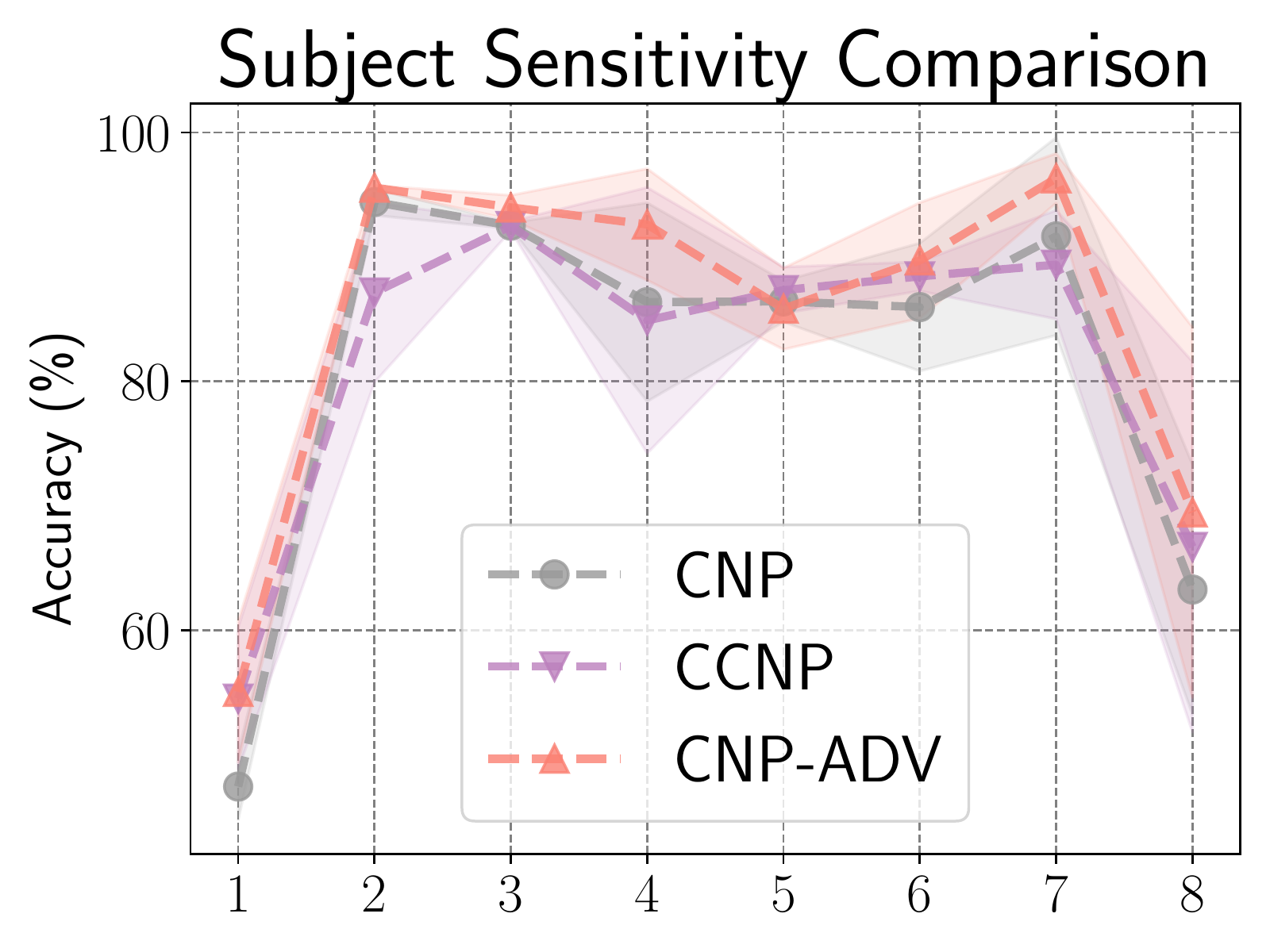}
    }
    \caption{The prediction accuracy of PAMAP activity recognition.}
    \vskip -0.3in
\end{figure}

\textbf{Results}.
Human activities manifest inter-subject variability.
That is, different patterns may emerge even when subjects perform the same cognitive task.
We thus report the average overall comparison~(\cref{fig:pamap_overall}) and subject-specific comparisons~(\cref{fig:pamap_sensitivity}).
While CCNP achieves higher average accuracy than CNP, it performs worse with a range of subjects.
In comparison, CNP-ADV consistently improves CNP with almost all subjects.
Even with subject~\#1 and~\#8 who have distinct activity patterns from others, CNP-ADV yields better prediction results than both CNP and CCNP.



\subsection{Complexity Overhead}
\label{subsec:complexity}
We have discussed that adversarial calibration benefits the expressivity of CNPs in terms of both generation and downstream tasks.
We also claim that it does not impose a heavy computational burden on the original CNP model.
It does not alter {\it how CNPs perform a complete training step} and {\it how CNPs make inference}.
The additional computation with respect to~\cref{eq:cnp_nce} is embarked by EBM that takes CNP predictions and true observations over all {\it target} points.
Still, the complexity is determined by the original CNP variant. 
For example, the asymptotic complexity of CNP is increased from $O(C + T)$ to $O(C + T + 2*T)$, but not enlarging its order.
We include more empirical comparisons of relative training time and memory consumptions in~\cref{appendix:complexity}.

\section{Discussion and Future Overlook}

\textbf{Summary}.
We presented an adversarial training scheme for calibrating the predictions of CNPs with NCE.
The expressivity of CNPs can be improved with a simple yet effective adversarial calibration process, by training a flexible unnormalized EBM that distinguishes true observations from ones generated by CNPs in the principle of NCE.
This provides a discriminative objective, by which CNPs must learn to fool EBM rather than simply following the pre-defined maximum likelihood estimate.
Our study demonstrated that CNPs would benefit more from NCE in adversarial training than from previous settings focusing on {\it context} representations, in terms of both generation and downstream tasks.

\textbf{Limitation and Future Extension}.
One limitation of this study is introduced by the pre-defined likelihood function of CNPs.
Despite adversarial training helping with predictions of CNPs, the function forms of factorized predictive distributions remain unchanged.
Combining this with tractable distribution transformation approaches, such as normalizing flows~(NFs), may provide a solution.
It is not trivial since NFs could challenge the consistency of CNPs as discussed in~\cite{markou2022practical}.
Another solution might be applying adversarial training to latent NPs, since they have a more flexible predictive space than CNPs.
However, it is unclear whether the NCE objective is directly applicable as latent NPs can only be estimated with approximated likelihoods.

\bibliography{reference}
\bibliographystyle{icml2023}

\newpage
\appendix
\onecolumn

\section{Consistency and Exchangeability}
\label{appendix:foundation}

Two foundation properties of CNPs are {\it exchangeability} and {\it consistency} under permutations and marginalization to satisfy the Kolmogorov Extension Theorem~\cite{oksendal2003stochastic}.
We next show that CNPs in adversarial training with NCE~(CNPs-ADV) do not violate such properties.

Suppose a data-generating function $f(\mbf{x})$ with a list of {\it target} input $\bs{x}_{1:T} = \{ \bs{x}_{t} \}_{t=1:T}$ have corresponding output $\bs{y}_{1:T} = \{ \bs{y}_{t} \}_{t=1:T}$.
Denote a random subset of {\it target} by $D_{C} := \{\bs{x}_{c}, \bs{y}_{c} \}_{c=1:C} \subset \{ \bs{x}_{t}, \bs{y}_{t} \}_{1:T}$.
Let $\rho_{T}$ be any permutation of $\{ 1, \dots, T \}$ and $\rho_{C}$ be any permutation of $\{ 1, \dots, C \}$.
Consider a CNP $\theta := \{ h(D_{C}), g(\bs{x}_t, p(f))\}$ and an EBM $\varphi(\cdot)$.

\begin{lemma}\label{appendix:lemma}
    For any covariate $\bs{x}_t$, the output of EBM $\varphi(\cdot)$ is an injective composition $\varphi \circ f: \mbb{R}^{d_{\mca{Y}}} \to \mbb{R}$ of a function $f(\mbf{x})$.

    \begin{proof}
        An EBM $\varphi(\cdot)$ takes either true observation $\bs{y}_{t} = f(\bs{x}_{t})$ or predictive observation $\dot{\bs{y}}_{t} = g(\bs{x}_t, p(f))$ as input.
        This holds in the fact that CNP collapses $p(f)$ into a deterministic tensor $\mbf{r}_{C}$ given $D_{C}$, and the decoder $g$ is a function of $\bs{x}_t$.
    \end{proof}

\end{lemma}

\begin{proposition}
    CNPs-ADV satisfy the exchangeability under any permutations.

    \begin{proof}
        The joint marginal distribution over $\bs{x}_{1:T}$ parameterized by a CNP $\theta$ is permutation-invariant to $\rho_{T}$ given by,
        \begin{equation}
            \begin{aligned}
                p(\bs{y}_{1}, \dots, \bs{y}_{T} | \bs{x}_{1}, \dots, \bs{x}_{T}) & = \int p(f) p(\bs{y}_{1}, \dots, \bs{y}_{T} | \bs{x}_{1}, \dots, \bs{x}_{T}, p(f)) \, \dev f \\
                                                          & \approx p(\bs{y}_{1}, \dots, \bs{y}_{T} | \bs{x}_{1}, \dots, \bs{x}_{T}, \bs{x}_{1},\dots, \bs{x}_{C}, \bs{y}_{1}, \dots, \bs{y}_{C}; \theta) \\
                                                          & = p(\bs{y}_{1}, \dots, \bs{y}_{T} | \bs{x}_{1}, \dots, \bs{x}_{T}, \mbf{r}_{C}; \theta) \\
                                                          & = \prod_{t=1}^{T} p(\bs{y}_{t} | \bs{x}_{t}, \mbf{r}_{C}; \theta)  \qquad \text{assume } p(\bs{y}_{t} | \bs{x}_{t} ; \theta) \indep p(\bs{y}_{t^{\prime}} | \bs{x}_{t} ; \theta) \mid \mbf{r}_{C}, \; \forall \; t \neq t^{\prime} \\
                                                          & = \prod_{t=1}^{T} p(\bs{y}_{\rho_{T}(t)} | \bs{x}_{\rho_{T}(t)}, \mbf{r}_{C}; \theta) \\
                                                          & = p(\bs{y}_{\rho_{T}(1)}, \dots, \bs{y}_{\rho_{T}(T)} | \bs{x}_{\rho_{T}(1)}, \dots, \bs{x}_{\rho_{T}(T)})
            \end{aligned}
        \end{equation}\label{eq:appendix_cnp}
        provided that
        \begin{equation}
            \begin{aligned}
                \mbf{r}_{C} & = \oplus_{c} \, h(\bs{x}_{c}, \bs{y}_{c}) = \oplus_{\rho_{c}(c)} \, h(\bs{x}_{\rho_{c}(c)}, \bs{y}_{\rho_{c}(c)}), \qquad \forall \, c \in 1, \dots, C
            \end{aligned}    
        \end{equation}
        such that $\mbf{r}_{C}$ is invariant to $\rho_{C}$ if $\oplus$ is a permutation-invariant aggregation operation, e.g., mean-pooling~\cite{garnelo2018conditional} and self-attention~\cite{kimanp18}.

        With~\cref{appendix:lemma}, the EBM outputs an injective $\varphi \circ f$ of $p(\bs{y}_t | \bs{x}_{t}, \mbf{r}_{C}, \theta)$ which is parameterized by $\dot{\bs{y}_{t}}$.
        The permutation $\rho_{T}$ on $\{1, \dots, T\}$ thus has an unique consequence on $\varphi (\dot{\bs{y}}_{\rho_{T}(t)})$
        \begin{equation}
            \varphi(\dot{\bs{y}}_{1}), \dots, \varphi(\dot{\bs{y}}_{T}) = \varphi(\dot{\bs{y}}_{\rho_{T}(1)}), \dots, \varphi(\dot{\bs{y}}_{\rho_{T}(T)})
        \end{equation}
        The same holds for $\varphi (\bs{y}_{\rho_{T}(t)})$.
    \end{proof}
\end{proposition}

\begin{proposition}
    CNPs-ADV satisfy the consistency under marginalization.

    \begin{proof}

        Given the conditional independence assumption within {\it target}, marginalizing out $\bs{y}_{1}, \dots, \bs{y}_{T - 1}$ results in the joint predictive distribution of $\bs{x}_{1}, \dots \bs{x}_{T}$ as
        \begin{equation}
            \begin{aligned}
                & \int \dots \int p(\bs{y}_{1}, \dots, \bs{y}_{T} | \bs{x}_{1}, \dots, \bs{x}_{T}, \mbf{r}_{C} ; \theta) \, \dev \bs{y}_{1} \dots \dev \bs{y}_{T-1} \\ 
                = & \int \dots \int p(\bs{y}_{T} | \bs{x}_{T}, \mbf{r}_{C}; \theta) p(\bs{y}_{1}, \dots, \bs{y}_{T-1} | \bs{x}_{1}, \dots, \bs{x}_{T-1}, \mbf{r}_{C} ; \theta) \, \dev \bs{y}_{1} \dots \dev \bs{y}_{T-1} \\
                = \, & p(\bs{y}_{T} | \bs{x}_{T}, \mbf{r}_{C}; \theta) \int p(\bs{y}_1 | \bs{x}_1, \mbf{r}_{C}; \theta) \, \dev \bs{y}_{1} \int \dots \int p(\bs{y}_{T - 1} | \bs{x}_{T - 1}, \mbf{r}_{C}; \theta) \, \dev \bs{y}_{T - 1} \\
                = \, & p(\bs{y}_{T} | \bs{x}_{T}, \mbf{r}_{C}; \theta)
            \end{aligned}
        \end{equation}

        As the density of $\dot{\bs{y}}_{T}$ remains unchanged depending on $\bs{x}_{t}$ and $\mbf{r}_{C}$.
        Resembling the {\it exchangeability} condition, the injective $\varphi \circ \dot{\bs{y}}_{T}$ does not violate the consistency requirement held by CNP.
    \end{proof}
\end{proposition}

\clearpage

\section{Additional Details for 1D Data.}
\label{appendix:sine_oscil_exp}

\subsection{Experiment Setup}

\subsubsection{Dataset Description}

\begin{figure}[!htbp]
    \subfigure[Examples of sine wave functions]{
        \includegraphics[width=0.45\columnwidth]{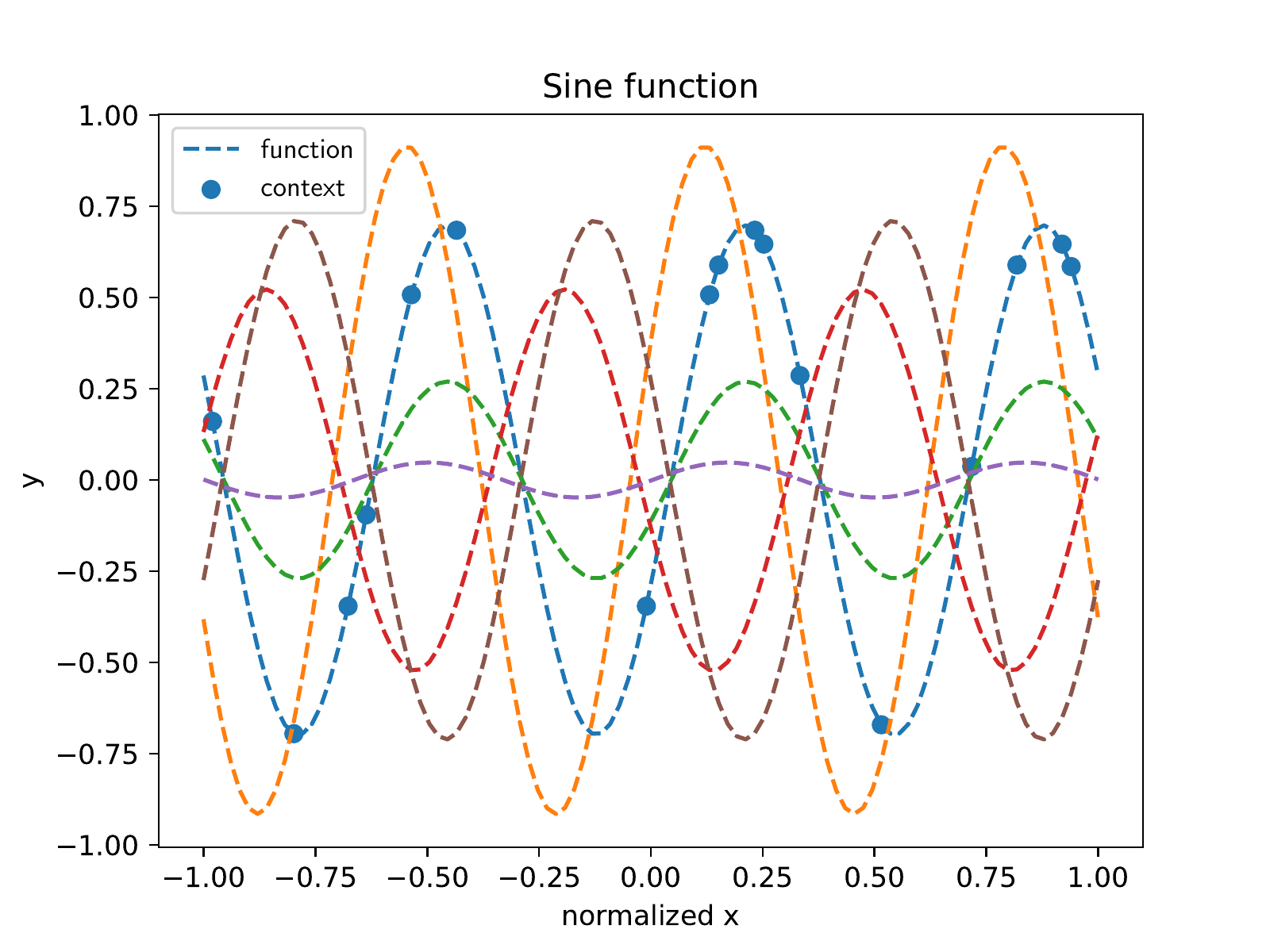}
    }
    \subfigure[Examples of oscillator functions]{
        \includegraphics[width=0.54\columnwidth]{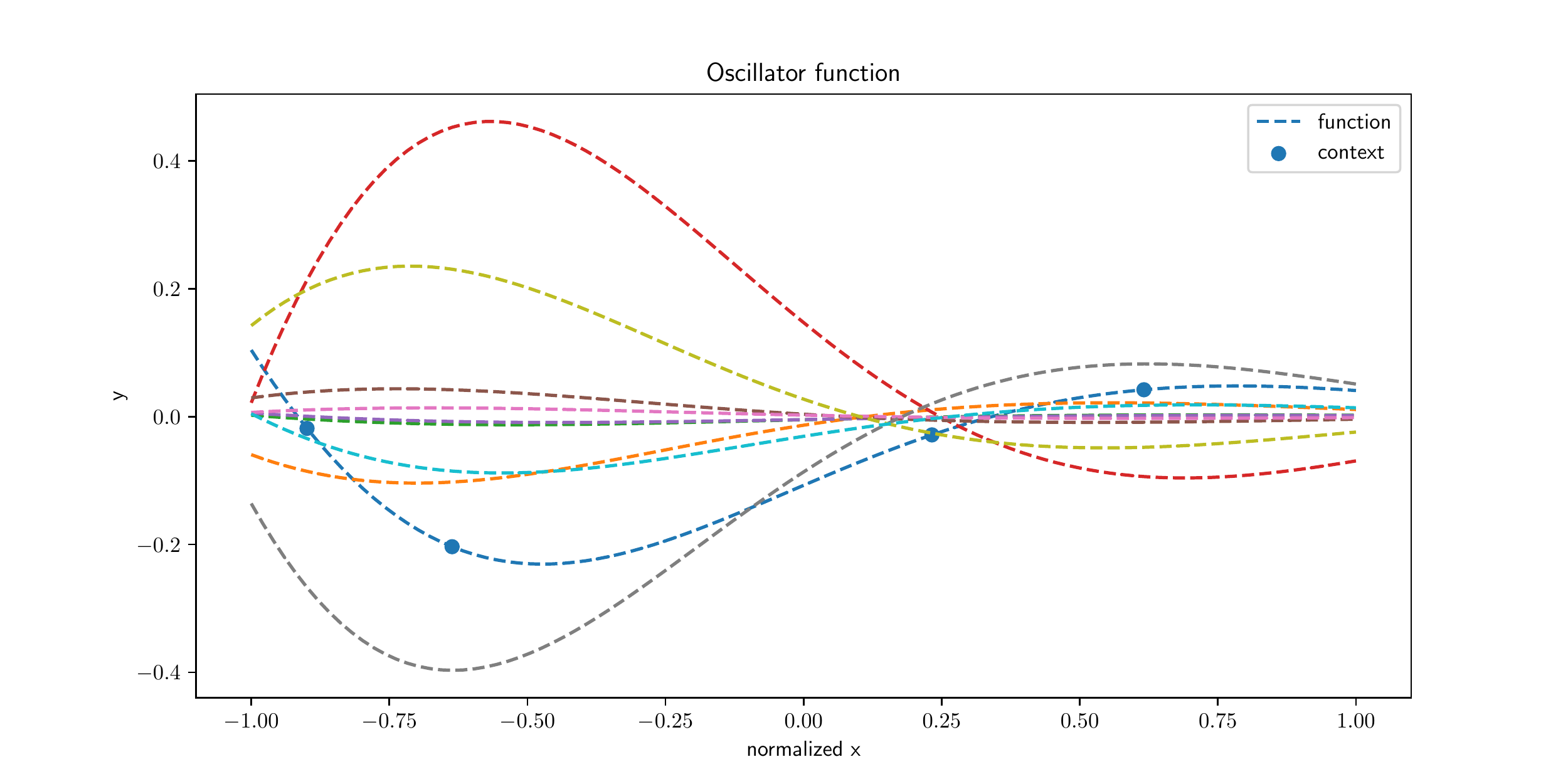}
    }
    \caption{Examples of 1D functions used in the experiment.}
\end{figure}

The sine wave functions are generated by
\begin{equation}
    y = a * \sin (x - b), \quad \text{with } a \sim \mca{U}[-1.0, 1.0], \text{ and } b \sim \mca{U}[-0.5, 0.5],
\end{equation}
where $\mbf{x}$ are evaluated from $[-3\pi, 3\pi]$.

The damped oscillator functions are generated by
\begin{equation}
    y = a * \exp(-0.5x) * \sin(x - b), \quad \text{with } a \sim \mca{U}[-3.0, 3.0], \text{ and } b \sim \mca{U}[-1.0, 1.0],
\end{equation}
where $\mbf{x}$ are evaluated from $[0, 5]$. 
Both function families have 100 points per instantiation.
The train/validation/test set has distinct 500 instantiations, respectively.

For Gaussian Processes~(GP) generated data, we adopt a commonly-used RBF kernel
\begin{equation}
    k\left(\mathbf{x}, \mathbf{x}^{\prime}\right)=\exp \left(-\frac{\left\|\mathbf{x}-\mathbf{x}^{\prime}\right\|^2}{2 \sigma^2}\right), \quad \text{with } \sigma = 0.2.
\end{equation}
where $\mbf{x}$ are evaluated from [-1., 1.] at 128 uniformly spacing covariates.
The train/set has 4,096 and 1,000 instantiations, respectively.

For each dataset, the covariate range $\mbf{x}$ is linearly scaled to $[-1, 1]$ regardless of their original lower and upper bound.

\subsubsection{Setup}

\textbf{CNPs Training}.
We use all the points of a function instantiation as the {\it target} size throughout this work.
We fix {\it context} size in evaluation to the upper bound of that in training.
For other datasets, these remain the same unless explicitly stated.

\begin{itemize}
    \item context size: $\mca{U}[0.04, 0.2]$ for sine and oscillator functions; $\mca{U}[0.1, 0.3]$ for GP-RBF functions.
    \item batch size: 16 for sine and oscillator functions; 128 for GP-RBF functions.
    \item episodes: 200 for sine and oscillator functions; 250 for GP-RBF functions.
    \item optimization: Adam with learning rate 1e-3. Weight decay 6e-5.
\end{itemize}

\textbf{CNPs-Adv Training}.
To calibrate CNPs predictions with NCE, we append an EBM following each original CNP variant.
Within each batch of $B \times T$ {\it target} points, EBM takes both $B \times T$ true observations and $B \times T$ predicted observations as input, and outputs a scalar value for each input.
The training of CNPs-Adv is in two stages.
In stage 1, we train CNPs as usual, until the validation loss does not decrease in 20 consecutive epochs.
In stage 2, we put EBM into joint training.
We obtain a value with respect to~\cref{eq:cnp_nce} and compute the prediction accuracy of EBM on the density ratio.
If the accuracy is greater than the threshold $\alpha$, then update CNP in the next batch, otherwise update EBM (with another optimizer).
\begin{itemize}
    \item iterative threshold $\alpha$: 0.6.
    \item optimization: Adam with learning rate 7e-4
\end{itemize}

\textbf{Downstream Task Training.}
We fix the parameters of pre-trained CNPs $\theta^{*}$, and use a prediction head $\phi(\mbf{r}_{C})$ on the {\it context} representation extracted by CNPs $\theta^{*}$.

The downstream regression task involves only oscillator functions.
We are tasked with predicting the amplitude $a$ or the shift $b$ given a particular function instantiation, i.e., $\phi: \mbb{R}^{d_{r}} \to \mbb{R}$.

\begin{itemize}
    \item context size: 0.2
    \item batch size: 32
    \item Episodes: 20
    \item Optimization: Adam with learning rate 5e-4. Weight decay 6e-5.
\end{itemize}

\subsection{Model Implementation}
We provide the implementations for 1D and basketball data in\footnote{\url{https://anonymous.4open.science/r/icml23_submission-D2BB}}\label{link:code}.

In practice, both covariates $\bs{x}$ and observations $\bs{y}$ can be multi-dimensional, i.e., $d_{\mca{X}} > 1, d_{\mca{Y}} > 1$.
Thus, we follow the interpretation of~\cite{mathieu2021contrastive} that applies a covariate network~(CovNet) and an observation network~(ObsNet) to {\it pre-transform} covariates and observations into features, such as a CNN to process image observations $\bs{y}$ with $d_{\mca{Y}} = 1024$.

\subsubsection{CNP}

\textbf{Sine and oscillator}.
\begin{itemize}
    \item CovNet: identity $\mbb{R} \to \mbb{R}$
    \item ObsNet: identity $\mbb{R} \to \mbb{R}$
    \item Encoder $h(\bs{x}, \bs{y})$: 2-(hidden)~layer MLP $\mbb{R}^{2} \to \mbb{R}^{64}$, with hidden size $[64, 64]$.
    \item Aggregator: mean-pooling
    \item Decoder $g(\bs{x}, \mbf{r}_{C})$: 2-layer MLP $\mbb{R}^{65} \to \mbb{R}$, with hidden size $[64, 64]$.
    \item Likelihood: diagonal Gaussian with event shape $1$.
\end{itemize}

\textbf{GP-RBF}.
\begin{itemize}
    \item CovNet: identity $\mbb{R} \to \mbb{R}$
    \item ObsNet: identity $\mbb{R} \to \mbb{R}$
    \item Encoder $h(\bs{x}, \bs{y})$: 3-(hidden)~layer MLP $\mbb{R}^{2} \to \mbb{R}^{128}$, with hidden size $[128, 128, 128]$.
    \item Aggregator: mean-pooling
    \item Decoder $g(\bs{x}, \mbf{r}_{C})$: 5-layer MLP $\mbb{R}^{129} \to \mbb{R}$, with hidden size $[128, 128, 128, 128, 128]$.
    \item Likelihood: diagonal Gaussian with event shape $1$.
\end{itemize}

\subsubsection{AttentiveCNP~(ACNP)}
\textbf{Sine and oscillator}.
\begin{itemize}
    \item CovNet: identity $\mbb{R} \to \mbb{R}$
    \item ObsNet: identity $\mbb{R} \to \mbb{R}$
    \item Encoder $h(\bs{x}, \bs{y})$: 2-(hidden)~layer MLP $\mbb{R}^{2} \to \mbb{R}^{64}$, with hidden size $[64, 64]$.
    \item Aggregator: identity\footnote{for 1D data, we follow the setting as in~\cite{kimanp18} where the authors do not use self-attention}
    \item Cross-Attention: multi-head scale-dot product attention $\mbb{R}^{64} \to \mbb{R}^{64}$, with 8 heads.
    \item Decoder $g(\bs{x}, \mbf{r}_{C})$: 2-layer MLP $\mbb{R}^{65} \to \mbb{R}$, with hidden size $[64, 64]$.
    \item Likelihood: diagonal Gaussian with event shape $1$.
\end{itemize}

\textbf{GP-RBF}.
\begin{itemize}
    \item CovNet: identity $\mbb{R} \to \mbb{R}$
    \item ObsNet: identity $\mbb{R} \to \mbb{R}$
    \item Encoder $h(\bs{x}, \bs{y})$: 3-(hidden)~layer MLP $\mbb{R}^{2} \to \mbb{R}^{128}$, with hidden size $[128, 128, 128]$.
    \item Aggregator: identity
    \item Cross-Attention: multi-head scale-dot product attention $\mbb{R}^{128} \to \mbb{R}^{128}$, with 8 heads.
    \item Decoder $g(\bs{x}, \mbf{r}_{C})$: 5-layer MLP $\mbb{R}^{129} \to \mbb{R}$, with hidden size $[128, 128, 128, 128, 128]$.
    \item Likelihood: diagonal Gaussian with event shape $1$.
\end{itemize}

\subsubsection{ContrastiveCNP~(CCNP)}
The difference between CCNP and CNP is that CCNP adopts an additional {\tt NCE Projector} on the batch of {\it context} sets, where the {\it context} points of each instantiation are split into two disjoint subsets. 
Then their mean representations are aligned with NCE - by considering the subsets from the same function as positive pairs, while contrasting with the subsets of other instantiations within the batch.

\textbf{Sine and oscillator}.
\begin{itemize}
    \item CovNet: identity $\mbb{R} \to \mbb{R}$
    \item ObsNet: identity $\mbb{R} \to \mbb{R}$
    \item Encoder $h(\bs{x}, \bs{y})$: 2-(hidden)~layer MLP $\mbb{R}^{2} \to \mbb{R}^{64}$, with hidden size $[64, 64]$.
    \item Aggregator: mean-pooling
    \item NCE Projector: 1-layer MLP $\mbb{R}^{64} \to \mbb{R}^{64}$, with hidden size $[64]$ and BatchNormalization.
    \item Decoder $g(\bs{x}, \mbf{r}_{C})$: 2-layer MLP $\mbb{R}^{65} \to \mbb{R}$, with hidden size $[64, 64]$.
    \item Likelihood: diagonal Gaussian with event shape $1$.
\end{itemize}

\textbf{GP-RBF}.
\begin{itemize}
    \item CovNet: identity $\mbb{R} \to \mbb{R}$
    \item ObsNet: identity $\mbb{R} \to \mbb{R}$
    \item Encoder $h(\bs{x}, \bs{y})$: 3-(hidden)~layer MLP $\mbb{R}^{2} \to \mbb{R}^{128}$, with hidden size $[128, 128, 128]$.
    \item Aggregator: mean-pooling
    \item NCE Projector: 1-layer MLP $\mbb{R}^{128} \to \mbb{R}^{64}$, with hidden size $[64]$ and BatchNormalization.
    \item Decoder $g(\bs{x}, \mbf{r}_{C})$: 5-layer MLP $\mbb{R}^{129} \to \mbb{R}$, with hidden size $[128, 128, 128, 128, 128]$.
    \item Likelihood: diagonal Gaussian with event shape $1$.
\end{itemize}

\subsubsection{Energy-Based Model~(EBM)}

\begin{itemize}
    \item EBM $\varphi(\bs{y})$: 1-layer MLP $\mbb{R} \to \mbb{R}$, with hidden size $[128]$
\end{itemize}


\subsubsection{Downstream Prediction Head}

\begin{itemize}
    \item prediction head $\phi(\mbf{r}_{C})$: 1-layer MLP $\mbb{R}^{64} \to \mbb{R}$, with hidden size $[128]$
\end{itemize}


\clearpage
\section{Additional Details for Basketball data.}
\label{appendix:basket}

\subsection{Experiment Setup}

\subsubsection{Dataset Description}

The basketball trajectory dataset is provided by STATS SportsVU tracking data from the 2012-2013 NBA season, taken from\footnote{https://github.com/ezhan94/multiagent-programmatic-supervision}.
Each sequence~(as a function instantiation) contains the trajectories of 10 players (of two teams) plus a basketball.
For each timestep, i.e., covariate $\bs{x}$, the observation has features $\bs{y} \in \mbb{R}^{22}$, in the order of the basketball, 5 offensive players and 5 defensive players.
Each trajectory describes the 2D positions of a player~(or the basketball) throughout 50 timesteps.

In total, there are 104,003 train sequences and other 13,464 test sequences.
We randomly pick 10,400 from train sequences for validation during training.

The covariate range $\mbf{x}$ is linearly scaled to $[-1, 1]$ from $[0, 49]$.
The observation values of each feature are z-scored.

\begin{figure}[!htbp]
    \begin{center}
        \includegraphics[width=0.65\columnwidth]{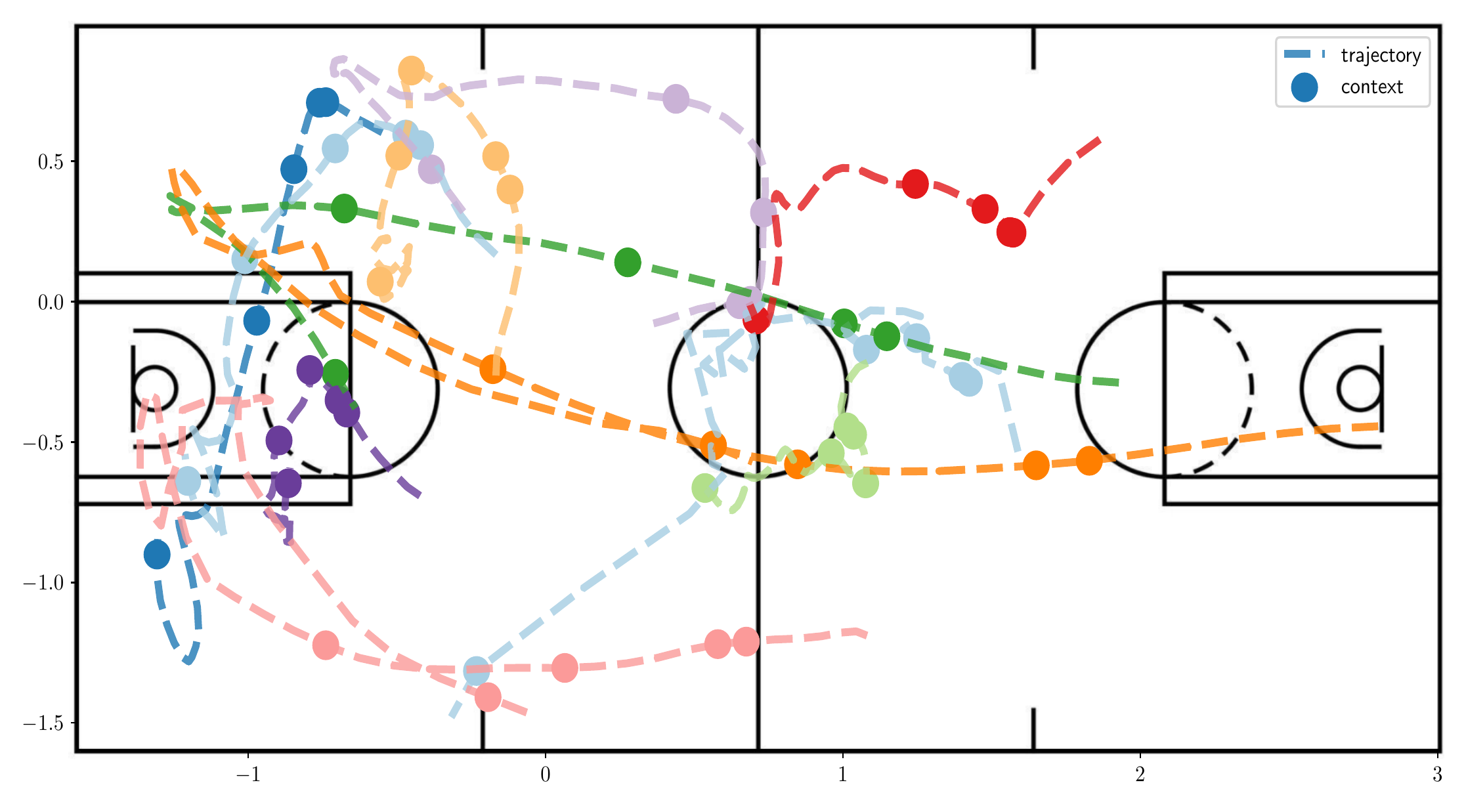}
        \caption{Illustrative example of a basketball in-game trajectories sequence. The values of coordinate are with z-score normalization.}    
    \end{center}
\end{figure}

\subsubsection{Setup}

\textbf{CNPs Training}.
We use all the points of a function instantiation as the {\it target} size throughout this work.
We fix {\it context} size in evaluation to the upper bound of that in training.
For other datasets, these remain the same unless explicitly stated.

\begin{itemize}
    \item context size: $\mca{U}[0.1, 0.5]$
    \item batch size: 128
    \item episodes: 200
    \item optimization: Adam with learning rate 1e-3. Weight decay 6e-5.
\end{itemize}

\textbf{CNPs-Adv Training}.
To calibrate CNPs predictions with NCE, we append an EBM following each original CNP variant.
Within each batch of $B \times T$ {\it target} points, EBM takes both $B \times T$ true observations and $B \times T$ predicted observations as input, and outputs a scalar value for each input.
The training of CNPs-Adv is in two stages.
In stage 1, we train CNPs as usual, until the validation loss does not decrease in 20 consecutive epochs.
In stage 2, we put EBM into joint training.
We obtain a value with respect to~\cref{eq:cnp_nce} and compute the prediction accuracy of EBM on the density ratio.
If the accuracy is greater than the threshold $\alpha$, then update CNP in the next batch, otherwise update EBM (with another optimizer).
\begin{itemize}
    \item iterative threshold $\alpha$: 0.6.
    \item optimization: Adam with learning rate 7e-4
\end{itemize}

\subsection{Model Implementation}

\subsubsection{CNP}

\begin{itemize}
    \item CovNet: identity $\mbb{R} \to \mbb{R}$
    \item ObsNet: MLP $\mbb{R}^{22} \to \mbb{R}^{128}$, without hidden layer.
    \item Encoder $h(\bs{x}, \bs{y})$: 3-(hidden)~layer MLP $\mbb{R}^{129} \to \mbb{R}^{128}$, with hidden size $[128, 128, 128]$.
    \item Aggregator: mean-pooling
    \item Decoder $g(\bs{x}, \mbf{r}_{C})$: 4-layer MLP $\mbb{R}^{129} \to \mbb{R}^{22}$, with hidden size $[128, 128, 128, 128]$.
    \item Likelihood: diagonal Gaussian with event shape $22$.
\end{itemize}

\subsubsection{AttentiveCNP~(ACNP)}
\begin{itemize}
    \item CovNet: identity $\mbb{R} \to \mbb{R}$
    \item ObsNet: MLP $\mbb{R}^{22} \to \mbb{R}^{128}$, without hidden layer.
    \item Encoder $h(\bs{x}, \bs{y})$: 3-(hidden)~layer MLP $\mbb{R}^{129} \to \mbb{R}^{128}$, with hidden size $[128, 128, 128]$.
    \item Aggregator: multi-head self-attention, with hidden size 256, 2 attention layers, 8 heads.
    \item Cross-Attention: multi-head scale-dot product attention $\mbb{R}^{64} \to \mbb{R}^{64}$, with 8 heads.
    \item Decoder $g(\bs{x}, \mbf{r}_{C})$: 4-layer MLP $\mbb{R}^{129} \to \mbb{R}^{22}$, with hidden size $[128, 128, 128, 128]$.
    \item Likelihood: diagonal Gaussian with event shape $22$.
\end{itemize}

\subsubsection{ContrastiveCNP~(CCNP)}

\begin{itemize}
    \item CovNet: identity $\mbb{R} \to \mbb{R}$
    \item ObsNet: MLP $\mbb{R}^{22} \to \mbb{R}^{128}$, without hidden layer.
    \item Encoder $h(\bs{x}, \bs{y})$: 3-(hidden)~layer MLP $\mbb{R}^{129} \to \mbb{R}^{128}$, with hidden size $[128, 128, 128]$.
    \item Aggregator: mean-pooling
    \item NCE Projector:  1-layer MLP $\mbb{R}^{128} \to \mbb{R}^{64}$, with hidden size $[64]$ and BatchNormalization.
    \item Decoder $g(\bs{x}, \mbf{r}_{C})$: 4-layer MLP $\mbb{R}^{129} \to \mbb{R}^{22}$, with hidden size $[128, 128, 128, 128]$.
    \item Likelihood: diagonal Gaussian with event shape $22$.
\end{itemize}

\subsubsection{Energy-Based Model~(EBM)}

\begin{itemize}
    \item EBM $\varphi(\bs{y})$: 1-layer MLP $\mbb{R}^{22} \to \mbb{R}$, with hidden size $[128]$
\end{itemize}


\clearpage

\section{Additional Details for Bouncing Balls data.}
\label{appendix:bball}

\subsection{Experiment Details}

\subsubsection{Dataset Description}

The high-dimensional synthesized image sequences are taken from\footnote{ https://github.com/cagatayyildiz/ODE2VAE}.
Each sequence describes the trajectories of three interacting balls inside a rectangular box.
Colliding with other balls or the ``wall'' would cause the balls to change directions.
All the models have no prior knowledge of the sequences, i.e., ball counts, initial velocity are unknown.
Each sequence has 50 frames of 1,024 pixels, i.e., $32 \times 32$ image.
That is, the observation $\bs{y} \in \mbb{R}^{1024}$.

There are 10,000 training sequences and 500 test sequences.
Similarly, we split 1000 validation sequences from the training set in training.

\begin{figure}[!htbp]
    \centering
    \includegraphics[width=0.8\columnwidth]{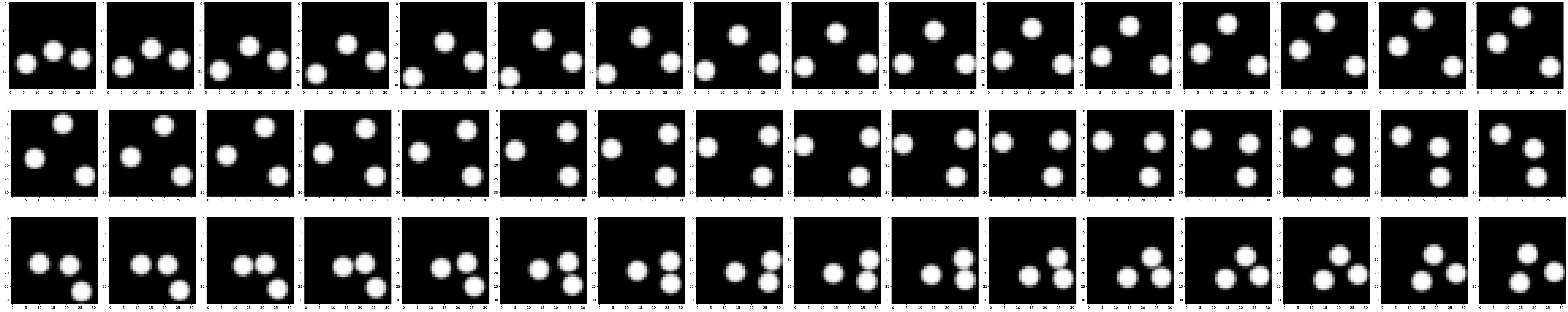}
    \caption{Examples of bouncing ball sequences.}
\end{figure}

\begin{figure}[!htbp]
    \centering
    \includegraphics[width=0.6\columnwidth]{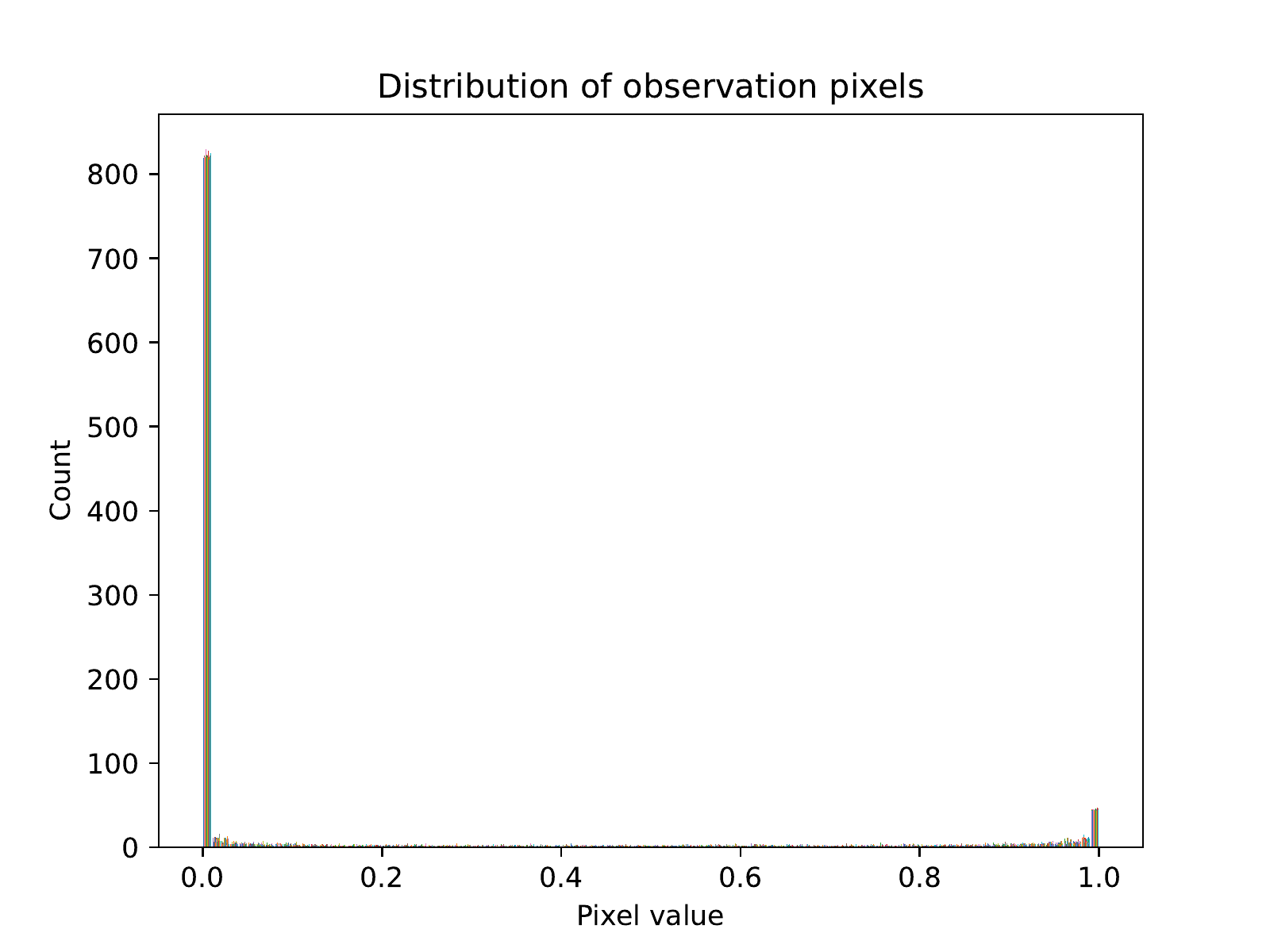}
    \caption{Distribution statistics of bouncing ball data observations.}
    \label{fig:appendix_bball_stat}
\end{figure}

\subsubsection{Setup}

\textbf{CNPs Training}.
We use all the points of a function instantiation as the {\it target} size throughout this work.
We fix {\it context} size in evaluation to the upper bound of that in training.
For other datasets, these remain the same unless explicitly stated.

\begin{itemize}
    \item context size: $\mca{U}[0.1, 0.5]$
    \item batch size: 128
    \item episodes: 200
    \item optimization: Adam with learning rate 1e-3. Weight decay 6e-5.
\end{itemize}

\textbf{CNPs-Adv Training}.
To calibrate CNPs predictions with NCE, we append an EBM following each original CNP variant.
Within each batch of $B \times T$ {\it target} points, EBM takes both $B \times T$ true observations and $B \times T$ predicted observations as input, and outputs a scalar value for each input.
The training of CNPs-Adv is in two stages.
In stage 1, we train CNPs as usual, until the validation loss does not decrease in 20 consecutive epochs.
In stage 2, we put EBM into joint training.
We obtain a value with respect to~\cref{eq:cnp_nce} and compute the prediction accuracy of EBM on the density ratio.
If the accuracy is greater than the threshold $\alpha$, then update CNP in the next batch, otherwise update EBM (with another optimizer).
\begin{itemize}
    \item iterative threshold $\alpha$: 0.6.
    \item optimization: Adam with learning rate 7e-4
\end{itemize}

\subsection{Model Implementation}

This dataset mainly differs from the others in the fact that the observations $\bs{y}$ are images.
We thus use a convolutional network for ObsNet, and a deconvolution network to recover the observation in the decoder $g$.
Specifically,

\subsubsection{CNP}
\begin{itemize}
    \item CovNet: identity $\mbb{R} \to \mbb{R}$
    \item ObsNet: CNN $\mbb{R}^{1024} \to \mbb{R}^{128}$, with 3 convolution blocks.
    \item Encoder $h(\bs{x}, \bs{y})$: 4-(hidden)~layer MLP $\mbb{R}^{129} \to \mbb{R}^{128}$, with hidden size $[128, 128, 128, 128]$.
    \item Aggregator: mean-pooling.
    \item Decoder $g(\bs{x}, \mbf{r}_{C})$: MLP $\Rightarrow$ DeconvNet $\mbb{R}^{129} \to \mbb{R}^{128} \to \mbb{R}^{1024}$, MLP has no hidden layer; DeConvNet has 4 deconvolution blocks.
    \item Likelihood: diagonal Bernoulli\footnote{because the pixel distribution of this dataset locates around 0 and 1, see~\cref{fig:appendix_bball_stat}} with event shape $1024$.
\end{itemize}

\subsubsection{ContrastiveCNP~(CCNP)}

\begin{itemize}
    \item CovNet: identity $\mbb{R} \to \mbb{R}$
    \item ObsNet: CNN $\mbb{R}^{1024} \to \mbb{R}^{128}$, with 3 convolution blocks.
    \item Encoder $h(\bs{x}, \bs{y})$: 4-(hidden)~layer MLP $\mbb{R}^{129} \to \mbb{R}^{128}$, with hidden size $[128, 128, 128, 128]$.
    \item Aggregator: mean-pooling
    \item NCE Projector:  1-layer MLP $\mbb{R}^{128} \to \mbb{R}^{64}$, with hidden size $[64]$ and BatchNormalization.
    \item Decoder $g(\bs{x}, \mbf{r}_{C})$: MLP $\Rightarrow$ DeconvNet $\mbb{R}^{129} \to \mbb{R}^{128} \to \mbb{R}^{1024}$, MLP has no hidden layer; DeConvNet has 4 deconvolution blocks.
    \item Likelihood: diagonal Bernoulli with event shape $1024$.
\end{itemize}

\subsubsection{Energy-Based Model~(EBM)}

\begin{itemize}
    \item EBM $\varphi(\bs{y})$: 1-layer MLP $\mbb{R}^{1024} \to \mbb{R}$, with hidden size $[128]$
\end{itemize}


\subsubsection{Additional Qualitative Results}

We include more qualitative reconstruction results of CNP, CCNP and CNP-ADV with respect to the ground-truth observations.
The examples demonstrate that CNP and CCNP may suffer from three balls gathered cases in particular. 
Conversely, CNP-ADV performs much better than the other two.

\begin{figure}
    \begin{center}
        \begin{small}
            \includegraphics[width=0.75\columnwidth]{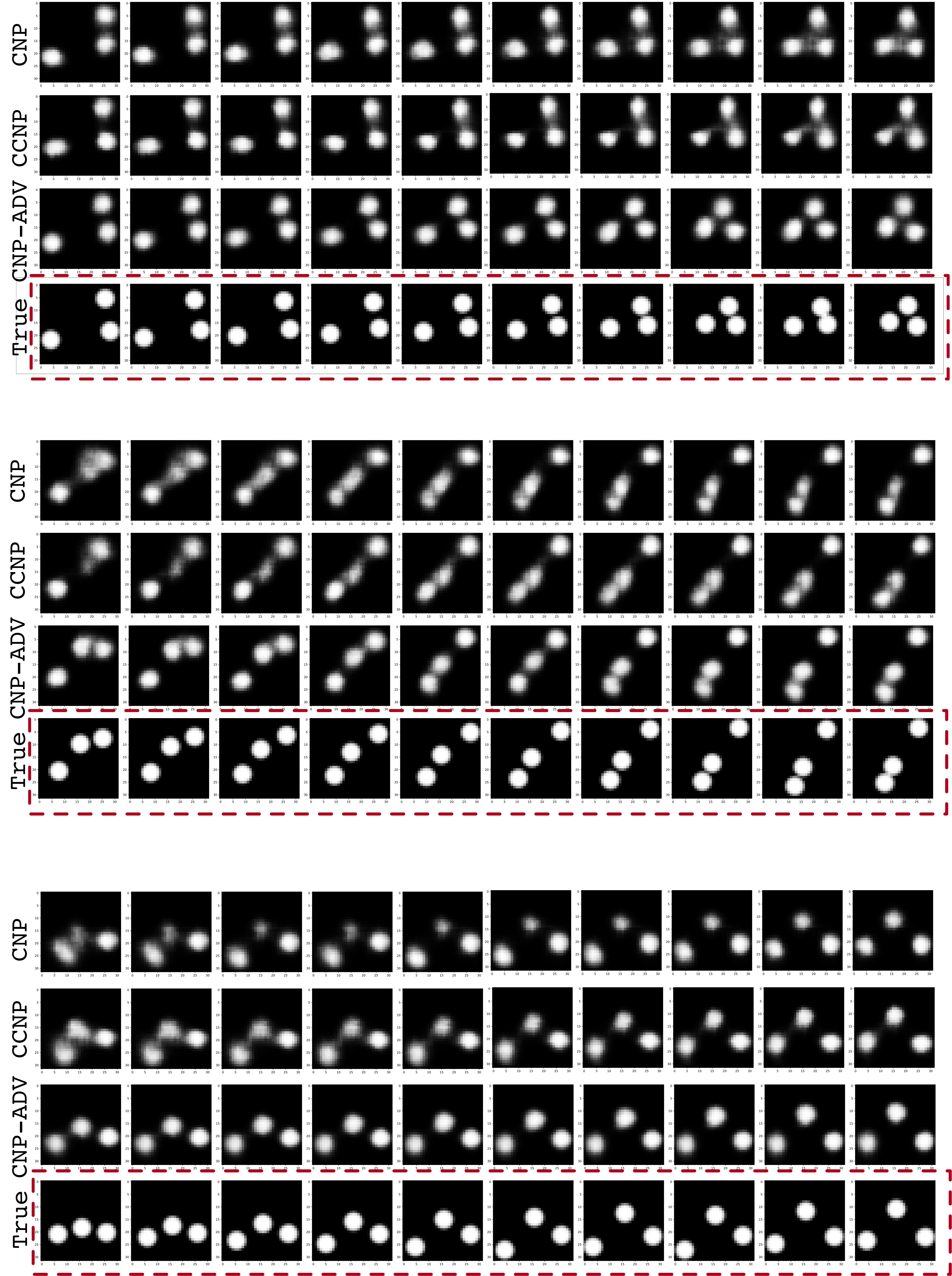}
            \caption{The reconstruction results with consecutive 10 timeframes for another three sequences.}
        \end{small}
    \end{center}

\end{figure}

\clearpage

\section{Additional Details for EEG data.}
\label{appendix:eeg}

\subsection{Experiment Setup}

\subsubsection{Dataset Description}
The EEG dataset we adopt in this study is a commonly-used one as in\footnote{https://physionet.org/content/eegmmidb/1.0.0/}.
This dataset is distributed under the license ODC-By\footnote{https://physionet.org/content/eegmmidb/view-license/1.0.0/}.
The EEG collection protocol provides 64-channel EEG recordings using the BCI2000 system.
There are in total 109 subjects involved in performing the Motor Imagery tasks.
We use the data from 105 subjects out of 109, except for subject \#88, \#89, \#92, \#100 due to their consecutively resting states.
For our experiments, we select the recordings upon performing two tasks:
\begin{itemize}
    \item task 0: open and close both fists or both feet;
    \item task 1: imaging opening and closing both fists or both feet.
\end{itemize}
We use the open-sourced EEG decoding toolset\footnote{https://mne.tools/stable/index.html} for data loading and preprocessing.
Each EEG trial is sliced into many 4-seconds clips corresponding to the task, with 160 samples per second.
Thus, each function instantiation in this dataset is of 640 timesteps with a label $l$ indicating the associated Motor Imagery task.

After gathering the data of 105 subject, we randomly sample 10\% clips for testing, 10\% clips for validation during training, and 80\% clips for model training.

The covariate range $\mbf{x}$ is linearly scaled to $[-1, 1]$ from $[0, 640]$.
The observation values of each feature are z-scored.

\subsubsection{Setup}

\textbf{CNPs Training}.
We use all the points of a function instantiation as the {\it target} size throughout this work.
We fix {\it context} size in evaluation to the upper bound of that in training.
For other datasets, these remain the same unless explicitly stated.

\begin{itemize}
    \item context size: $\mca{U}[0.5, 0.8]$
    \item batch size: 128
    \item episodes: 200
    \item optimization: Adam with learning rate 1e-3. Weight decay 6e-5.
\end{itemize}

\textbf{CNPs-Adv Training}.
To calibrate CNPs predictions with NCE, we append an EBM following each original CNP variant.
Within each batch of $B \times T$ {\it target} points, EBM takes both $B \times T$ true observations and $B \times T$ predicted observations as input, and outputs a scalar value for each input.
The training of CNPs-Adv is in two stages.
In stage 1, we train CNPs as usual, until the validation loss does not decrease in 20 consecutive epochs.
In stage 2, we put EBM into joint training.
We obtain a value with respect to~\cref{eq:cnp_nce} and compute the prediction accuracy of EBM on the density ratio.
If the accuracy is greater than the threshold $\alpha$, then update CNP in the next batch, otherwise update EBM (with another optimizer).
\begin{itemize}
    \item iterative threshold $\alpha$: 0.6.
    \item optimization: Adam with learning rate 7e-4
\end{itemize}

\textbf{Downstream Task Training.}
We fix the parameters of pre-trained CNPs $\theta^{*}$, and use a prediction head $\phi(\mbf{r}_{C})$ on the {\it context} representation extracted by CNPs $\theta^{*}$.

We are tasked with predicting the label $l$ given a particular EEG clip, i.e., $\phi: \mbb{R}^{d_{r}} \to \mbb{R}^{2}$.
The objective is cross entropy.

\begin{itemize}
    \item context size: 0.8
    \item batch size: 64
    \item Episodes: 10
    \item Optimization: Adam with learning rate 5e-4. Weight decay 6e-5.
\end{itemize}

\subsection{Model Implementation}

\subsubsection{CNP}

\begin{itemize}
    \item CovNet: identity $\mbb{R} \to \mbb{R}$
    \item ObsNet: 2-layer MLP $\mbb{R}^{64} \to \mbb{R}^{128}$, with hidden size [128, 128].
    \item Encoder $h(\bs{x}, \bs{y})$: 3-(hidden)~layer MLP $\mbb{R}^{129} \to \mbb{R}^{128}$, with hidden size $[128, 128, 128]$.
    \item Aggregator: mean-pooling
    \item Decoder $g(\bs{x}, \mbf{r}_{C})$: 4-layer MLP $\mbb{R}^{129} \to \mbb{R}^{64}$, with hidden size $[128, 128, 128, 128]$.
    \item Likelihood: diagonal Gaussian with event shape $64$.
\end{itemize}

\textbf{Model Structure}

\subsubsection{ContrastiveCNP~(CCNP)}

\begin{itemize}
    \item CovNet: identity $\mbb{R} \to \mbb{R}$
    \item ObsNet: 2-layer MLP $\mbb{R}^{64} \to \mbb{R}^{128}$, with hidden size [128, 128].
    \item Encoder $h(\bs{x}, \bs{y})$: 3-(hidden)~layer MLP $\mbb{R}^{129} \to \mbb{R}^{128}$, with hidden size $[128, 128, 128]$.
    \item Aggregator: mean-pooling
    \item NCE Projector:  1-layer MLP $\mbb{R}^{128} \to \mbb{R}^{64}$, with hidden size $[64]$ and BatchNormalization.
    \item Decoder $g(\bs{x}, \mbf{r}_{C})$: 4-layer MLP $\mbb{R}^{129} \to \mbb{R}^{64}$, with hidden size $[128, 128, 128, 128]$.
    \item Likelihood: diagonal Gaussian with event shape $64$.
\end{itemize}

\textbf{Model Structure}

\subsubsection{Energy-Based Model~(EBM)}

\begin{itemize}
    \item EBM $\varphi(\bs{y})$: 1-layer MLP $\mbb{R}^{64} \to \mbb{R}$, with hidden size $[128]$
\end{itemize}


\subsubsection{Downstream Prediction Head}

\begin{itemize}
    \item prediction head $\phi(\mbf{r}_{C})$: 1-layer MLP $\mbb{R}^{128} \to \mbb{R}^{2}$, with hidden size $[128]$
\end{itemize}


\clearpage

\section{Additional Details for PAMAP2 data.}
\label{appendix:pamap}

\subsection{Experiment Setup}

\subsubsection{Dataset Description}
The PAMAP2 dataset is a sensor-based physical activity monitoring dataset\footnote{https://archive.ics.uci.edu/ml/datasets/pamap2+physical+activity+monitoring}, commonly studied in the research area of data mining and ubiquitous computing.
The dataset is distributed under CC BY 4.0 license\footnote{https://archive-beta.ics.uci.edu/dataset/231/pamap2+physical+activity+monitoring}.
There were 18 subjects participating in the experiments, wearing 3 inertial measurement units~(IMUs) on hand, chest, and ankle, with a sampling frequency of 100Hz, and performing the required daily physical activities.
We use the data from 8 subjects~(id: 1-8) who performed the following 4 activities.
\begin{itemize}
    \item activity 0: lying
    \item activity 1: sitting
    \item activity 2: standing
    \item activity 3: walking
\end{itemize}

Each IMU records 17 features, with the first being temperature and the last four invalid orientation\footnote{each IMU is of temperature (1), 3D-acceleration (2-4), 3D-acceleration (5-7), 3D-acceleration (8-10) 3D-magnetometer (11-13), invalid orientation (14-17).} were excluded.
This results in $3 \times 12$ effective features per timestep, i.e., $\bs{y} \in \mbb{R}^{36}$.

Meanwhile, the raw signals contain missing values.
We apply sliding window on the raw signals with window size 50~(\~1 second) and keep consecutive timesteps only as each windowed data.

With each run, we use the data of 1 subject for testing (in both pre-training CNPs and downstream classification task), 1 subject for validation, while using the remaining 6 subjects for model training.

The covariate range $\mbf{x}$ is linearly scaled to $[-1, 1]$ from $[0, 49]$.
The observation values of each feature are z-scored.

\subsubsection{Setup}

\textbf{CNPs Training}.
We use all the points of a function instantiation as the {\it target} size throughout this work.
We fix {\it context} size in evaluation to the upper bound of that in training.
For other datasets, these remain the same unless explicitly stated.

\begin{itemize}
    \item context size: $\mca{U}[0.5, 0.8]$
    \item batch size: 128
    \item episodes: 200
    \item optimization: Adam with learning rate 1e-3. Weight decay 6e-5.
\end{itemize}

\textbf{CNPs-Adv Training}.
To calibrate CNPs predictions with NCE, we append an EBM following each original CNP variant.
Within each batch of $B \times T$ {\it target} points, EBM takes both $B \times T$ true observations and $B \times T$ predicted observations as input, and outputs a scalar value for each input.
The training of CNPs-Adv is in two stages.
In stage 1, we train CNPs as usual, until the validation loss does not decrease in 20 consecutive epochs.
In stage 2, we put EBM into joint training.
We obtain a value with respect to~\cref{eq:cnp_nce} and compute the prediction accuracy of EBM on the density ratio.
If the accuracy is greater than the threshold $\alpha$, then update CNP in the next batch, otherwise update EBM (with another optimizer).
\begin{itemize}
    \item iterative threshold $\alpha$: 0.6.
    \item optimization: Adam with learning rate 7e-4
\end{itemize}

\textbf{Downstream Task Training.}
We fix the parameters of pre-trained CNPs $\theta^{*}$, and use a prediction head $\phi(\mbf{r}_{C})$ on the {\it context} representation extracted by CNPs $\theta^{*}$.

We are tasked with predicting the label $l$ given a particular signal window, i.e., $\phi: \mbb{R}^{d_{r}} \to \mbb{R}^{4}$.
The objective is cross entropy.

\begin{itemize}
    \item context size: 0.8
    \item batch size: 64
    \item Episodes: 10
    \item Optimization: Adam with learning rate 5e-4. Weight decay 6e-5.
\end{itemize}

\subsection{Model Implementation}

\subsubsection{CNP}

\begin{itemize}
    \item CovNet: identity $\mbb{R} \to \mbb{R}$
    \item ObsNet: 2-layer MLP $\mbb{R}^{36} \to \mbb{R}^{128}$, with hidden size [128, 128].
    \item Encoder $h(\bs{x}, \bs{y})$: 3-(hidden)~layer MLP $\mbb{R}^{129} \to \mbb{R}^{128}$, with hidden size $[128, 128, 128]$.
    \item Aggregator: mean-pooling
    \item Decoder $g(\bs{x}, \mbf{r}_{C})$: 4-layer MLP $\mbb{R}^{129} \to \mbb{R}^{36}$, with hidden size $[128, 128, 128, 128]$.
    \item Likelihood: diagonal Gaussian with event shape $36$.
\end{itemize}

\textbf{Model Structure}

\subsubsection{ContrastiveCNP~(CCNP)}

\begin{itemize}
    \item CovNet: identity $\mbb{R} \to \mbb{R}$
    \item ObsNet: 2-layer MLP $\mbb{R}^{36} \to \mbb{R}^{128}$, with hidden size [128, 128].
    \item Encoder $h(\bs{x}, \bs{y})$: 3-(hidden)~layer MLP $\mbb{R}^{129} \to \mbb{R}^{128}$, with hidden size $[128, 128, 128]$.
    \item Aggregator: mean-pooling
    \item NCE Projector:  1-layer MLP $\mbb{R}^{128} \to \mbb{R}^{64}$, with hidden size $[64]$ and BatchNormalization.
    \item Decoder $g(\bs{x}, \mbf{r}_{C})$: 4-layer MLP $\mbb{R}^{129} \to \mbb{R}^{36}$, with hidden size $[128, 128, 128, 128]$.
    \item Likelihood: diagonal Gaussian with event shape $36$.
\end{itemize}

\subsubsection{Energy-Based Model~(EBM)}

\begin{itemize}
    \item EBM $\varphi(\bs{y})$: 1-layer MLP $\mbb{R}^{36} \to \mbb{R}$, with hidden size $[128]$
\end{itemize}


\subsubsection{Downstream Prediction Head}

\begin{itemize}
    \item prediction head $\phi(\mbf{r}_{C})$: 1-layer MLP $\mbb{R}^{128} \to \mbb{R}^{4}$, with hidden size $[128]$
\end{itemize}


\clearpage

\subsection{Subject-specific classification results}

We provide the exact classification accuracy of each subject.
The reported results are averaged over three different runs~(seeds).
\begin{table}[!htbp]
    \begin{center}
        \begin{small}
            \resizebox{\columnwidth}{!}{%
            \begin{tabular}{lcccccccc}
                \toprule
                        & 1 & 2 & 3 & 4 & 5 & 6 & 7 & 8 \\
                \midrule
                CNP     & 47.45 $\pm$ 2.69 & 94.41 $\pm$ 1.07 & 92.51 $\pm$ 0.24 & 86.37 $\pm$ 1.65 & 86.43 $\pm$ 1.65 & 85.97 $\pm$ 5.14 & 91.65 $\pm$ 7.94 & 63.28 $\pm$ 10.04 \\
                CCNP    & 54.51 $\pm$ 5.71 & 87.14 $\pm$ 7.22 & 92.48 $\pm$ 0.38 & 84.90 $\pm$ 10.68 & 87.33 $\pm$ 1.86 & 88.44 $\pm$ 1.17 & 89.38 $\pm$ 4.37 & 66.7 $\pm$ 14.89 \\
                CNP-ADV & \textbf{55.06} $\pm$ 5.69 & \textbf{95.57} $\pm$ 0.17 & \textbf{93.98} $\pm$ 0.97 & \textbf{92.61} $\pm$ 4.48 & 85.84 $\pm$ 3.27 & \textbf{89.72} $\pm$ 4.64 & \textbf{96.32} $\pm$ 2.0 & \textbf{69.48} $\pm$ 2.74 \\
                \bottomrule
            \end{tabular}
            }
        \end{small}
    \end{center}
    \caption{Subject-specific classification accuracy over 8 subjects}
    \label{tab:my_label}
\end{table}
The PAMAP dataset manifests inter-subject variability, especially on subjects \#1, and \#8.
When handling these two subjects, the {\it context} representation obtained from the training data of other subjects seems not sufficiently effective for all three models.
Still, the CNP-ADV appears to be the best performer among the three.
Meanwhile, CNP-ADV consistently demonstrates higher accuracy for most subjects except for subject \#5.

\clearpage

\section{Balancing NCE with MLE in adversarial training.}
\label{appendix:obj_ratio}

We have discussed a practical objective for updating CNPs in stage 2 - using a weighted combination of MLE and NCE.
Below we showcase our empirical findings in varying the value of $\beta$, i.e., the combination coefficient, with 1D sine data.

\begin{figure}[!htbp]
    \vskip 0.15in
    \centering
    \includegraphics[width=0.8\columnwidth]{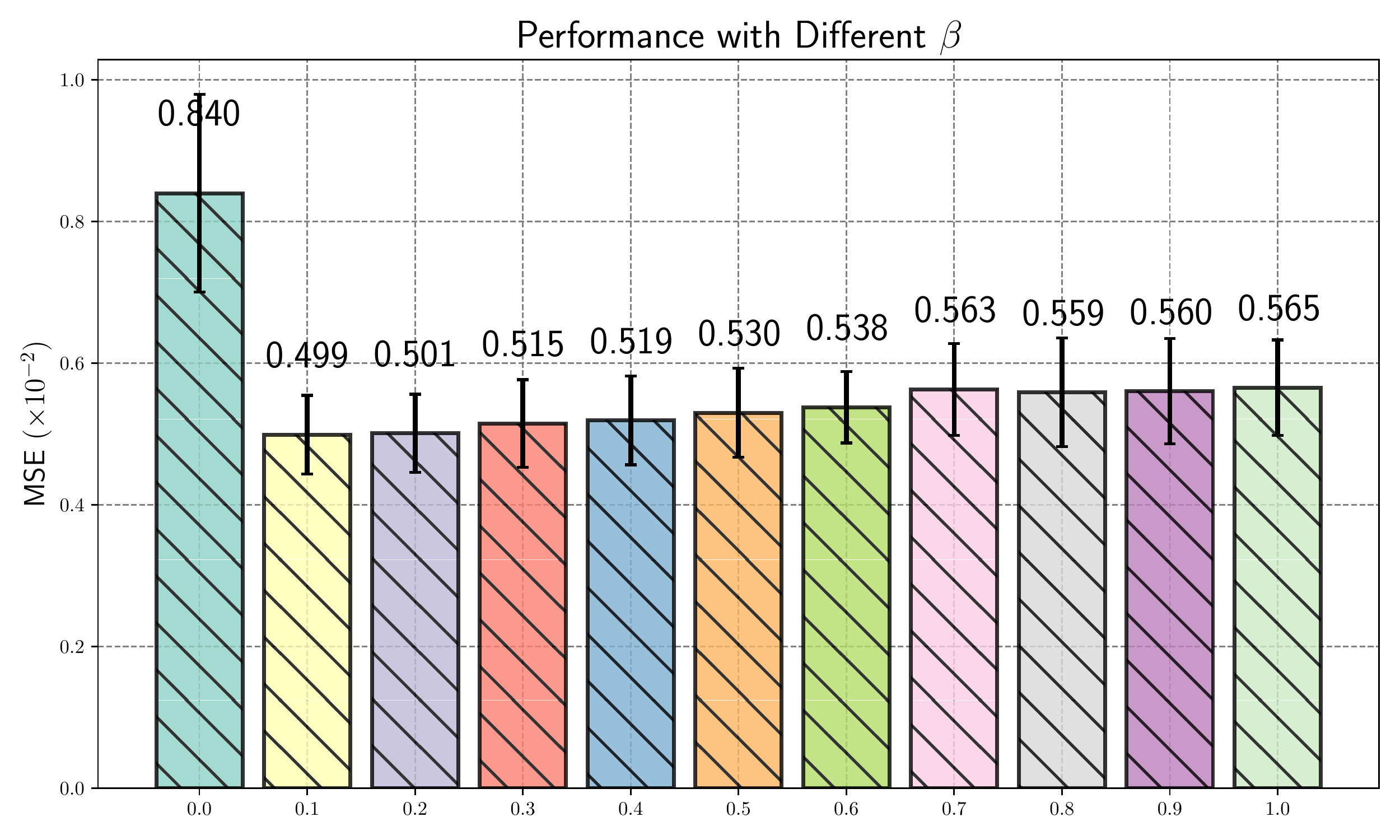}
    \caption{The comparison of different tradeoff coefficient $\beta$ values in MLE-NCE joint training objectives on 1D sine wave data. Results are obtained with three random seeds~(same for all $\beta$)}
    \label{fig:beta}
\end{figure}

As~\cref{fig:beta} shows, varying the value of $\beta$ gives different reconstruction performances with all the other settings kept the same.
When $\beta = 1.0$, the update to CNP $\theta$ is fully determined by the value of NCE.
In fact, this corresponds to the lowest results among all different $\beta$ values.
This may suggest the design choice of $\beta$ in practice.

We also find that $\beta=0.1$ works best for all the experiments involved in this work.
Note that this combination is used only for updating CNP.
As part of the iterative optimization, the EBM parameters $\varphi$ are solely optimized with respect to the NCE value.

\clearpage

\section{Computational Complexity Comparison.}
\label{appendix:complexity}

In~\cref{subsec:complexity}, we have discussed the complexity overhead on CNPs introduced by adversarial training, taking CNP for example, from $O(C + T)$ to $O(C + T + 2T)$, where $C$ is the number {\it context} points and $T$ is the number of {\it target} points.
The additional $2T$ account for the computation of EBM with $T$ true observations and $T$ CNP predictions.

We now include an empirical comparison between CNP and CNP-ADV with all the datasets studied in this paper.
The comparison is based on elapsed time difference with one epoch.
We perform the {\it relative} comparison, i.e., the elapsed time of CNP for each dataset is set to 1.
Then we can compare how much longer it will take for CNP-ADV than for CNP.

The ratio of overhead elapsed time depends on the specific network design~(i.e., encoder $h$, decoder $g$) of CNPs, as well as the size of dataset~(i.e., the number of functions).

\begin{figure}[!htbp]
    \vskip 0.15in
    \centering
    \includegraphics[width=0.9\columnwidth]{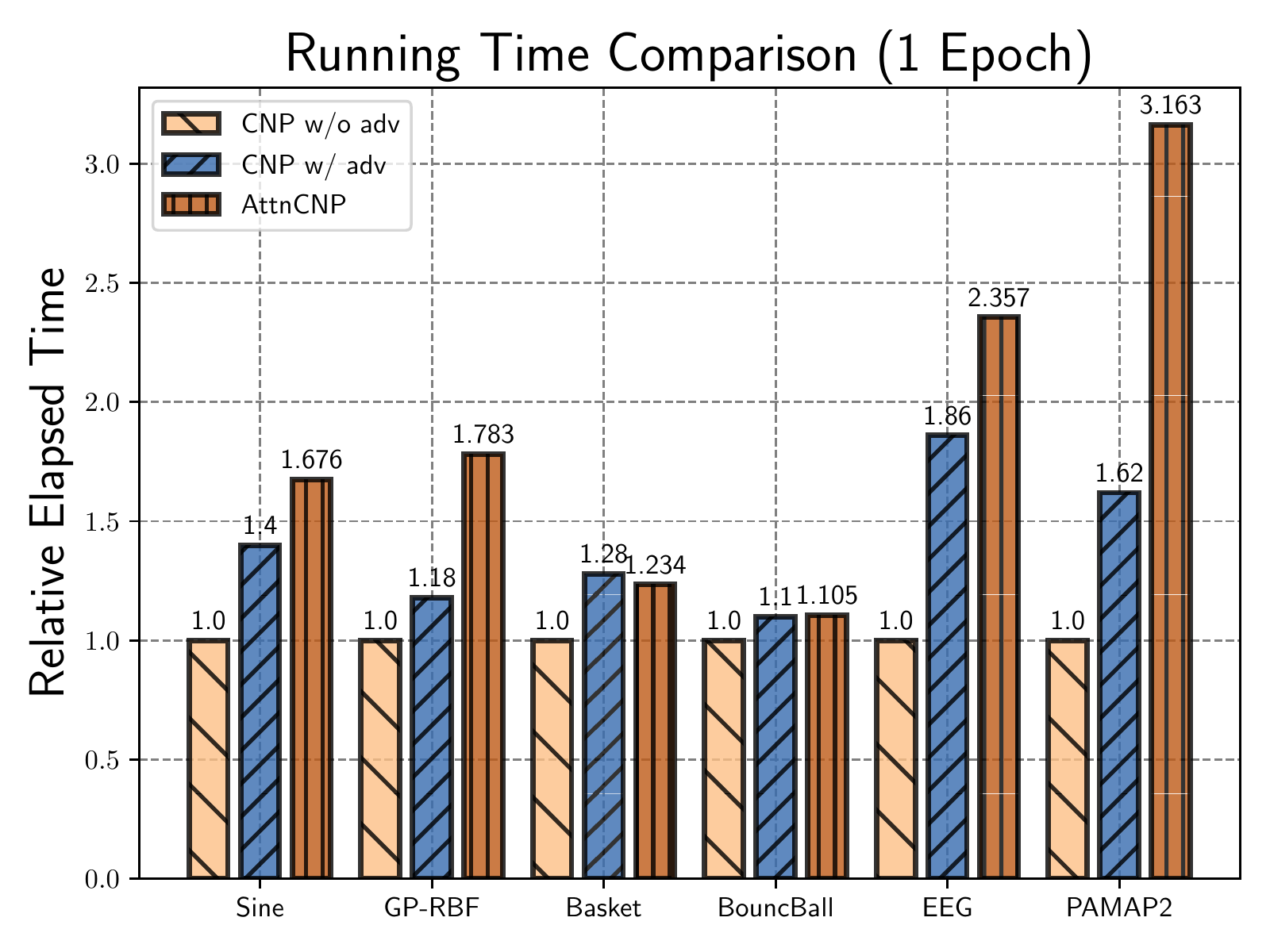}
    \caption{The comparison of {\it relative} elapsed time without/with adversarial training of running for 1 epoch.}
    \label{fig:my_label}
\end{figure}

We also include the relative running time between CNP and AttentiveCNP~\cite{kimanp18} for a clearer comparison.
The complexity of AttentiveCNP~(ACNP) is of $O(C( C + T))$ because self-attention for every {\it context} and cross-attention for every {\it target}.

\end{document}